%% file: journal.tex
\documentclass[lettersize,journal]{IEEEtran}
\usepackage{amsmath,amsfonts}
\usepackage{algorithmic}
\usepackage{algorithm}
\usepackage{array}
\usepackage[caption=false,font=normalsize,labelfont=sf,textfont=sf]{subfig}
\usepackage{textcomp}
\usepackage{stfloats}
\usepackage{url}
\usepackage{verbatim}
\usepackage{graphicx}
\usepackage{cite}
\usepackage{booktabs}
\usepackage{multirow}
\usepackage{tabularx}
\usepackage{siunitx}
\usepackage[section]{placeins}
\usepackage{xcolor}
\usepackage{tikz}
\usetikzlibrary{arrows.meta,positioning}
\begin{document}

\title{Timed Rule-Based Supervision of an End-to-End Autonomous Parking Policy}

\author{Kejia Gao, Liguo Zhou*, Lei Yu, Alois Knoll,~\IEEEmembership{Fellow,~IEEE,}
}

\markboth{Manuscript}%
{Gao \MakeLowercase{\textit{et al.}}: Timed Rule-Based Supervision of an End-to-End Autonomous Parking Policy}


\maketitle

\begin{abstract} 
We study whether a manually specified runtime supervisor can correct recurring failures of an existing end-to-end parking policy in a fixed CARLA parking lot. The vision-based Transformer architecture is inherited from Yang et al.; our contribution is a timed, rule-based Parametric Safety Shield (PSS) applied to its control outputs. The PSS uses hand-calibrated speed, position, and duration thresholds to intervene in observed failure modes, including boundary exits, delayed braking, and stalled or oscillatory control. In the reported closed-loop evaluation, 16 held-out target slots and six initial poses are each evaluated in four rounds (384 attempts per configuration). Target success increases from 327/384 (85.16\%) for the retrained policy to 375/384 (97.66\%) with the PSS; mean position and orientation errors among successful attempts are 0.21~m and 0.33$^\circ$. These results show an improvement within this simulator setup. The repeated attempts share one map, vehicle, and sensor configuration, and the PSS uses simulator world coordinates; thus the results do not establish generalization to other lots or real vehicles, or a formal safety guarantee.
\end{abstract}


\section{Introduction}

\IEEEPARstart{E}{nd-to-end} learning maps sensor inputs to control outputs and has been studied for autonomous parking. Parking is a structured, low-speed task, but collisions and incorrect stops remain possible. We examine a narrow question: whether a manually calibrated supervisor can correct repeatable failure modes of an existing policy in a known simulated lot.

In closed loop, isolated control errors can accumulate into parking failures. The E2E Parking Dataset~\cite{gao2025e2eparkingdatasetopen} supports reproduction of the E2E Parking model~\cite{yang2024e2e} in CARLA. The retrained model achieved 85.16\% target success in the reported evaluation. Inspection of its failures in the study lot found boundary exits, alternating brake and throttle after stopping, and delayed braking near obstacles. These observations motivated the specific interventions studied here; they do not establish a generalization gap outside the tested setup.

We attach a Parametric Safety Shield (PSS), a hand-designed timed supervisor, to the output of the existing Transformer policy. The supervisor overwrites selected commands when its speed, position, and timing guards fire. This is an engineering extension of the published policy rather than a new neural architecture or a proof of safety. In the evaluated CARLA configuration, target success rises from 85.16\% to 97.66\%, with mean position and orientation errors of 0.21~m and 0.33$^\circ$ among successful attempts.

The main contributions of this paper are as follows:
\begin{itemize}
    \item We specify a timed, hand-calibrated output supervisor for the previously published E2E Parking policy and document its guards, overrides, and parameter values.
    \item We report a closed-loop comparison with the same retrained policy without supervision, plus component ablations, within one CARLA lot. The main observed change is 48 additional target successes in 384 repeated attempts.
    \item We release the implementation at \url{https://github.com/KejiaGao/e2e-parking-carla-timed-automaton} to support replication and evaluation in further settings.
\end{itemize}

\section{Related Works}
We focus on parking policies and the perception and rule-based components relevant to the proposed supervisor.

\subsection{Transformer Model}

Transformers support multi-modal feature fusion in driving systems, including TransFuser~\cite{chitta2022transfuser}, InterFuser~\cite{shao2023safety}, and UniAD~\cite{hu2023planning}. These works concern broader driving tasks and do not test the parking output supervisor studied here.

Transformer models are also studied for parking. ParkPredict+ \cite{shen2022parkpredict+} forecasts waypoints. E2E Parking \cite{yang2024e2e} provides the camera-to-control Transformer architecture and CARLA evaluation pipeline used as the nominal policy in this study. Our earlier E2E Parking Dataset work \cite{gao2025e2eparkingdatasetopen} released data and reported 85.16\% target success after retraining that architecture. ParkingE2E \cite{li2024parkinge2e} instead predicts future waypoints from camera images and reports real-vehicle evaluation in four garages. HOPE \cite{jiang2025hope} combines reinforcement learning with Reeds--Shepp curves for parking path planning and evaluates diverse scenario difficulty, including real-world tests. These recent methods address different decision layers and evaluation environments; their published success rates are not directly comparable with our CARLA control-output intervention.

\subsection{BEV Representation}

The inherited E2E Parking policy projects camera features into bird's-eye view (BEV) using Lift-Splat-Shoot~\cite{philion2020lift}. BEVDepth~\cite{li2023bevdepth} demonstrates explicit depth supervision for multi-view BEV estimation. Our work retains the published parking network's BEV path and changes only its executed control commands.

\subsection{Rule-based Methods}

Rule-based planning has been combined with learned components in driving~\cite{bouchard2022rule,likmeta2020combining}. HOPE~\cite{jiang2025hope} is especially relevant to parking because it combines a learned planner with geometric Reeds--Shepp structure. Our PSS is narrower: it applies fixed, manually calibrated guards to the outputs of an unchanged learned policy.

Our scope is a runtime correction of the published E2E Parking controller's outputs. This differs from changing the learned policy, but it retains the familiar limitations of manually designed guards and calibrated thresholds.

\section{Method}
\begin{figure*}[t]
\centering
\begin{tikzpicture}[>=Latex, node distance=8mm, every node/.style={font=\small}, block/.style={draw,rounded corners,align=center,minimum height=12mm,text width=31mm}, inherited/.style={block,fill=blue!5,draw=blue!60}, added/.style={block,fill=red!5,draw=red!65}]
\node[inherited] (obs) {Cameras, vehicle state, target slot};
\node[inherited,right=of obs] (net) {Inherited E2E Parking Transformer};
\node[added,right=of net] (pss) {Proposed timed PSS\\guards and overrides};
\node[block,right=of pss] (plant) {Vehicle in CARLA\\executed command};
\draw[->] (obs) -- (net);
\draw[->] (net) -- (pss);
\draw[->] (pss) -- (plant);
\draw[->] (plant.south) -- ++(0,-5mm) -| node[pos=.25,below,font=\scriptsize] {next observation and simulator position/velocity for PSS} (obs.south);
\end{tikzpicture}
\caption{Complete closed-loop architecture. Blue boxes are inherited from E2E Parking~\cite{yang2024e2e}; the red box is the hand-designed component introduced here. The PSS uses CARLA world coordinates and vehicle velocity in addition to the nominal command.}
\label{fig:hybrid}
\end{figure*}
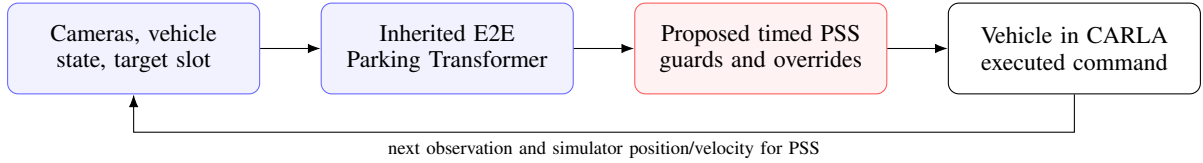

We retain the network architecture of Yang et al.~\cite{yang2024e2e} as the nominal controller. The new component in this paper is the post-inference PSS. Figure~\ref{fig:hybrid} shows the complete signal path and separates inherited and added components.

\subsection{Problem Definition}

The system is intended to output accurate control signal of parking by using input data of sensors. The model is trained in a supervised learning framework with expert demonstration dataset $\mathcal{D} = \{\tau^n \mid n \in \{1,2,\dots,N\}\}$, where each of the N trajectories $\tau_n$ consists of sequential sensor data and control signals: $\tau^n = \left\{ \left( \mathcal{X}_t^n, \mathcal{C}_t^n \right) \right\}_{t=1}^{T^n}$. $\mathcal{X}_t^n$ includes RGB images of four cameras, target slot coordinates and ego vehicle's dynamic state in the $n$th trajectory at time step $t$, while $\mathcal{C}_t^n$ consists of values of throttle, brake, steering and gear state in the $n$th trajectory at time step $t$.

The training objective is defined as follows:
\begin{equation}
\arg\min_{\pi} \mathbb{E}_{(\mathcal{X},\mathcal{C})\sim\mathcal{D}} \left[ \mathcal{L} \left( \mathcal{C}, \pi(\mathcal{X}) \right) \right]
\end{equation}

where $\pi$ denotes the control policy that maps observations to control outputs and $\mathcal{L}$ is the loss function.

\subsection{Inputs and Outputs}

\textbf{Inputs:} The model's input includes camera images, motion states (velocity and acceleration), and the target slot coordinates. Four surrounding cameras (front, left, right, rear) are deployed to capture the environmental images. The target slot is initially set by the user and then continuously tracked through segmentation task.

\textbf{Outputs:} The model forecasts control signals for the next four steps at a frequency of \SI{10}{\hertz}. The outputs consist of normalized acceleration (\(\text{acc}_t \in [-1,1]\)), steering angle (\(\text{steer}_t \in [-1,1]\)), and gear selection (\(\text{gear}_t \in \{0,1\}\)). Inspired by the approach of Pix2Seq \cite{chen2021pix2seq},  we regard control prediction as a language modeling task, where signals are tokenized and predicted sequentially. Acceleration and steering are discretized into 201 natural numbers \([0,200]\), where 0 signifies full brake/left and 200 indicating full throttle/right, and 100 represents no brake, throttle or steering. For gear state, 0 means forward gear and 200 indicates reverse gear. The sequence format is:

\begin{equation}
\begin{aligned}
S = [\text{BOS}, c_0, c_1, c_2, c_3, \text{EOS}] \\ 
c_n = [\text{acc}_n, \text{steer}_n, \text{gear}_n].
\end{aligned}
\end{equation}

{
\subsection{End-to-End Neural Network Architecture}

The following BEV generation, feature fusion, and control prediction stages describe the inherited E2E Parking network of Yang et al.~\cite{yang2024e2e}; they are reproduced for implementation context. Figure~\ref{fig:network} depicts that network alone, not the complete method in Fig.~\ref{fig:hybrid}.

\begin{figure*}[t]
\centering
\begin{tikzpicture}[>=Latex, node distance=5mm, every node/.style={font=\small}, stage/.style={draw,rounded corners,align=center,minimum height=14mm,text width=27mm,fill=blue!5,draw=blue!60}]
\node[stage] (images) {Four camera images};
\node[stage,right=of images] (bev) {EfficientNet-B4 and LSS\\BEV features};
\node[stage,right=of bev] (fusion) {Target slot and ego motion fusion};
\node[stage,right=of fusion] (transformer) {Transformer encoder and decoder};
\node[stage,right=of transformer] (control) {Control token sequence};
\draw[->] (images) -- (bev);
\draw[->] (bev) -- (fusion);
\draw[->] (fusion) -- (transformer);
\draw[->] (transformer) -- (control);
\end{tikzpicture}
\caption{Schematic of the inherited E2E Parking nominal policy, adapted from Yang et al.~\cite{yang2024e2e}. Auxiliary depth and segmentation losses are described in the text. The new PSS is shown separately in Fig.~\ref{fig:hybrid}.}
\label{fig:network}
\end{figure*}
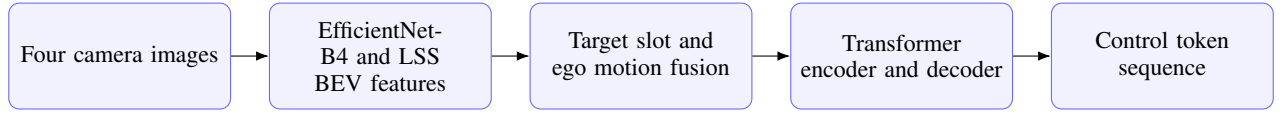

We use this vision-based imitation learning network as the nominal policy. The description below summarizes its operations rather than claiming a new neural architecture.

\subsubsection{BEV Feature Generation}
Given multi-view surrounding images $\mathcal{I} = \{I_n\}_{n=1}^N$ ($I_n \in \mathbb{R}^{3 \times H \times W}$), an EfficientNet-B4 backbone extracts multi-view 2D feature maps $F_{\text{img}, n} = \text{Backbone}(I_n) \in \mathbb{R}^{C_i \times H_i \times W_i}$. Following the LSS framework \cite{philion2020lift}, a depth classification head predicts discrete distributions $D_n \in \mathbb{R}^{D \times H_i \times W_i}$ over $D$ bins. A 3D context frustum $F_{\text{frus}, n}$ is generated via the outer product:
\begin{equation}
    F_{\text{frus}, n} = D_n \otimes F_{\text{img}, n} \in \mathbb{R}^{D \times C_i \times H_i \times W_i}
\end{equation}
Using camera intrinsics and extrinsics, these frustums are back-projected into a 3D voxel space and aggregated along the height axis via sum-pooling, yielding the BEV feature map $F_{\text{bev}} \in \mathbb{R}^{C_b \times X_b \times Y_b}$.

\subsubsection{Multi-Modal Feature Fusion}
The visual BEV map is unified with spatial intent and kinematics. The target parking slot is encoded into a binary grid $F_{\text{target}} \in \{0, 1\}^{1 \times X_b \times Y_b}$ and channel-wise concatenated with $F_{\text{bev}}$ to form $F_{\text{bev}}' = [F_{\text{bev}} \,;\, F_{\text{target}}] \in \mathbb{R}^{(C_b+1) \times X_b \times Y_b}$. 

To compress redundancy, $F_{\text{bev}}'$ is passed through a ResNet18 network and flattened into a visual sequence $F_{\text{bev\_seq}} \in \mathbb{R}^{C_{b\_ds} \times L}$, where $C_{b\_ds}$ is the downsampled channel size and $L$ is the token length. Concurrently, the vehicle's kinematic vector $F_{\text{ego0}} \in \mathbb{R}^{1 \times N_{\text{ego}}}$ is projected by an MLP into motion features $F_{\text{ego}} \in \mathbb{R}^{2 \times L}$. The final representation $F_{\text{fuse}}$ is unified via a multi-head self-attention Transformer encoder to capture global spatial correlations:
\begin{equation}
    F_{\text{fuse}} = \text{TransformerEncoder}\left( \begin{bmatrix} F_{\text{bev\_seq}} \\ F_{\text{ego}} \end{bmatrix} \right) \in \mathbb{R}^{(C_{b\_ds}+2) \times L}
\end{equation}

\subsubsection{Auto-Regressive Control Prediction}
A Transformer decoder maps $F_{\text{fuse}}$ into actuator profiles autoregressively. The cross-attention mechanism utilizes $F_{\text{fuse}}$ to populate the Key ($K$) and Value ($V$) matrices: $K = W_K F_{\text{fuse}}, V = W_V F_{\text{fuse}}$, while an empty query sequence with positional embeddings serves as the Query ($Q$). The decoder iteratively generates control tokens, which a linear layer decodes into raw vehicle commands (throttle, brake, steer, gear) prior to entering the PSS layer.

\subsection{Multi-Task Learning Loss Functions}
The network is optimized via a multi-task loss function:
\begin{equation}
    \mathcal{L}_{\text{total}} = \lambda_1 \mathcal{L}_{\text{ctrl}} + \lambda_2 \mathcal{L}_{\text{seg}} + \lambda_3 \mathcal{L}_{\text{depth}}
\end{equation}

\noindent \textbf{a. Control Loss ($\mathcal{L}_{\text{ctrl}}$):} A discrete cross-entropy loss enforces exact matching with expert demonstrations: $\mathcal{L}_{\text{ctrl}} = -\sum_{t=1}^{T} \sum_{k} y_{t,k} \log \hat{y}_{t,k}$, where $y_{t,k}$ and $\hat{y}_{t,k}$ are the ground-truth and predicted probabilities for the $k$-th control bin at step $t$.

\noindent \textbf{b. Segmentation Loss ($\mathcal{L}_{\text{seg}}$):} A semantic head categorizes the BEV grid into \textit{vehicle}, \textit{target slot}, and \textit{background} using categorical cross-entropy. During inference, closed-loop tracking is maintained by extracting the spatial centroid $(\bar{x}, \bar{y})$ of the predicted target slot to serve as $F_{\text{target}}$ for the next step. 

\noindent \textbf{c. Auxiliary Depth Loss ($\mathcal{L}_{\text{depth}}$):} Inspired by BEVDepth \cite{li2023bevdepth}, discrete depth estimations $D_n$ are supervised via binary cross-entropy using one-hot encoded LiDAR points projected onto camera planes, substantially enhancing spatial interpretability.
}

{
\subsection{Timed Rule-Based Output Supervision via PSS}

The PSS is a manually designed state machine with duration counters. Its guards were selected from observed failure cases in the study lot, and its numerical values were calibrated for that lot and vehicle. When a guard fires, the PSS replaces part or all of the nominal command. Figure~\ref{fig:PSS} shows the state topology. We use the term ``shield'' for this runtime supervisor; no invariant proof, reachability analysis, or formal safety guarantee is provided.

\begin{figure}[htbp]
    \centering
    \includegraphics[width=\linewidth]{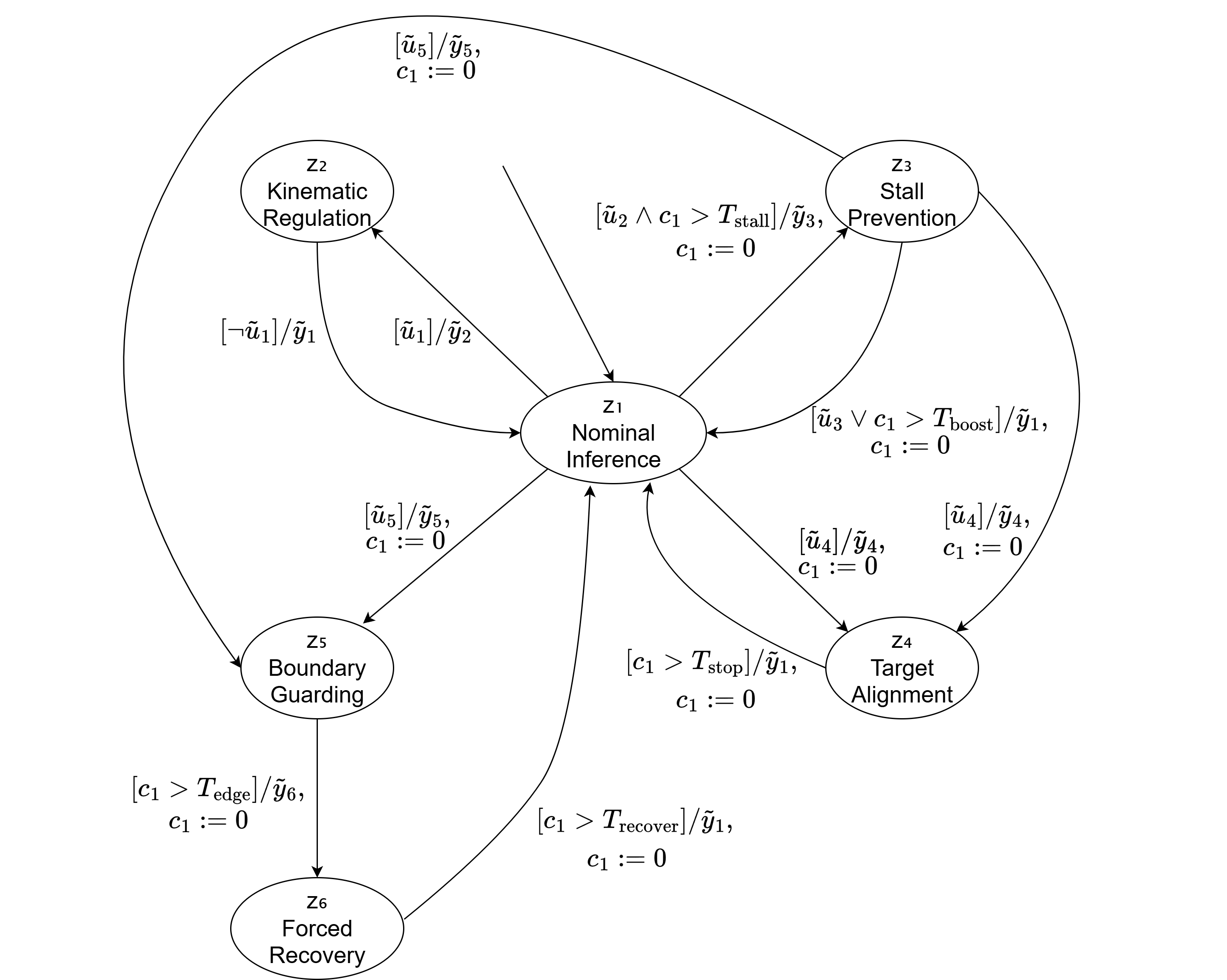}
    \caption{Topology of the PSS structured as a finite-state supervisor.}
    \label{fig:PSS}
\end{figure}

The supervisor runs at $10\ \mathrm{Hz}$ ($\Delta t = 0.1\ \mathrm{s}$). A discrete counter $c_1$ tracks time spent in selected modes. Its states ($\mathcal{Z}$), guards ($\mathcal{U}$), and control overrides ($\mathcal{Y}$) are listed below to make the implemented intervention reproducible. These conditions do not define a proved safe set.

\subsubsection{Formal Formulation of the Safety Shield}

\noindent \textbf{a. Operational States ($\mathcal{Z}$)}
\begin{itemize}
    \item $z_1: \text{Nominal Inference}$ --- The default state in which the Transformer command passes through; it does not imply that the state is safe.
    \item $z_2: \text{Kinematic Regulation}$ --- Activated when the vehicle velocity violates the safe kinematic upper bound, overriding the throttle to enforce immediate deceleration.
    \item $z_3: \text{Stall Prevention (Boost)}$ --- Addresses the low-velocity dead-zone problem common in imitation learning by injecting a persistent micro-torque to overcome static friction.
    \item $z_4: \text{Target Alignment (Stop)}$ --- Triggered upon reaching the precise terminal stopping threshold along the primary parking axis.
    \item $z_5: \text{Boundary Guarding}$ --- Initiated near the configured lot boundary to reduce observed run-off events.
    \item $z_6: \text{Forced Recovery}$ --- A structured, time-bounded reverse maneuver triggered sequentially after $z_5$ to guide the vehicle back into the safe operational region.
\end{itemize}

\begin{table}[htbp] 
\caption{Parameter Calibration of the PSS}
\label{tab:parameter_calibration}
\centering
\footnotesize
\begin{tabular}{llc}
\toprule
\textbf{Parameter} & \textbf{Description} & \textbf{Calibrated Value} \\
\midrule
$v_{\text{limit}}$    & Kinematic velocity upper bound             & $7.5\ \mathrm{km/h}$ \\
$v_{\text{stall}}$    & Low-velocity stall threshold               & $2.0\ \mathrm{km/h}$ \\
$\delta_{x}$          & Longitudinal alignment tolerance           & $0.075\ \mathrm{m}$ \\
$Y_{\text{boundary}}$ & Spatial boundary of the parking lot        & $-239.4\ \mathrm{m}$ \\
$v_{y,\text{th}}$     & Lateral velocity run-off threshold         & $-0.01\ \mathrm{km/h}$ \\
$\epsilon_{\text{f}}$ & Numerical deadband floating zero  & $1 \times 10^{-5}$ \\
\midrule
$\alpha_{\text{rev\_th}}$ & Throttle lower bound in reverse mode   & $0.11$ \\
$\alpha_{\text{boost}}$   & Stall prevention micro-throttle input  & $0.3$ \\
$\alpha_{\text{recover}}$ & Forced recovery throttle input         & $0.3$ \\
$P_{\text{stop}}$         & Terminal alignment braking input       & $0.9$ \\
$P_{\text{edge}}$         & Boundary guarding braking input        & $0.3$ \\
\midrule
$T_{\text{stall}}$    & Maximum allowable stall duration           & $21\ \text{steps}$ \\
$T_{\text{boost}}$    & Boost torque injection duration            & $10\ \text{steps}$ \\
$T_{\text{stop}}$     & Hold duration for terminal stop            & $21\ \text{steps}$ \\
$T_{\text{edge}}$     & Reaction buffer at spatial edge            & $4\ \text{steps}$ \\
$T_{\text{recover}}$  & Forced recovery phase duration             & $25\ \text{steps}$ \\
\bottomrule
\end{tabular}
\end{table}

\noindent \textbf{b. Parametric Guard Conditions ($\mathcal{U}$)}
\begin{align}
    \tilde{u}_1 \!: & \quad \left| v_x \right| \ge \left| v_{\text{limit}} \right| \\
    \tilde{u}_2 \!: & \quad \begin{aligned}[t]
        \Big( &(\neg R \;\wedge\; u_{\text{th}} < \epsilon_{\text{f}}) \;\vee\; (R \;\wedge\; u_{\text{th}} < \alpha_{\text{rev\_th}}) \Big) \\
        &\wedge\; u_{\text{br}} < \epsilon_{\text{f}} \;\wedge\; \left| v_x \right| < \left| v_{\text{stall}} \right|
    \end{aligned} \\
    \tilde{u}_3 \!: & \quad u_{\text{br}} > \epsilon_{\text{f}} \\
    \tilde{u}_4 \!: & \quad \left| x - x_{\text{goal}} \right| < \delta_{x} \\
    \tilde{u}_5 \!: & \quad \neg R \;\wedge\; \left| y \right| > \left| Y_{\text{boundary}} \right| \;\wedge\; \left| v_y \right| > \left| v_{y,\text{th}} \right|
    \label{eq:guards}
\end{align}
where $v_x$ and $v_y$ denote the vehicle longitudinal and lateral velocities; $x$ and $y$ represent the spatial coordinates; $R \in \{\text{True}, \text{False}\}$ is the reverse flag; $u_{\text{th}}$ and $u_{\text{br}}$ are the throttle and brake control commands, respectively.

\noindent \textbf{c. Control Output Mappings ($\mathcal{Y}$)}
\begin{itemize}
    \item $\tilde{y}_1 \!:~ \text{No intervention (Pass-through neural policy outputs)}$
    \item $\tilde{y}_2 \!:~ \texttt{throttle} = 0$
    \item $\tilde{y}_3 \!:~ \texttt{throttle} = \alpha_{\text{boost}}$
    \item $\tilde{y}_4 \!:~ \texttt{throttle} = 0, \; \texttt{brake} = P_{\text{stop}}, \; \texttt{steer} = 0, \; \texttt{reverse} = \text{True}$
    \item $\tilde{y}_5 \!:~ \texttt{throttle} = 0, \; \texttt{brake} = P_{\text{edge}}, \; \texttt{reverse} = \text{True}$
    \item $\tilde{y}_6 \!:~ \texttt{throttle} = \alpha_{\text{recover}}, \; \texttt{brake} = 0, \; \texttt{steer} = \texttt{steer\_set}, \; \texttt{reverse} = \text{True}$
\end{itemize}
where the recovery steering angle $\texttt{steer\_set}$ is adaptively computed based on the spatial error vector:
\begin{equation}
\texttt{steer\_set} = 
\begin{cases} 
\theta_{\text{max}}, & \text{if } x < x_{\text{goal}} \\ 
-\theta_{\text{max}}, & \text{otherwise} 
\end{cases}
\end{equation}

\subsubsection{System Calibration and Generalization}
Table~\ref{tab:parameter_calibration} lists the 16 hand-chosen values used in this study. Parameterization makes these choices visible, but it does not make the state-transition logic environment independent. In particular, $Y_{\text{boundary}}$ is an absolute coordinate in the CARLA map, the terminal guard uses the goal's $x$ coordinate, and the overrides use fixed actuator values and durations. A different layout, vehicle, localization system, or sensor suite may require changes to both parameters and guards. We have not measured the amount of retuning required.
}

\section{Experiments}\label{Experiments}

\subsection{Experimental Setup}\label{setup}

The network architecture keeps the same as the original one and hyperparameters are slightly different. The network was trained on 4 NVIDIA RTX 4090 (total VRAM: 96 GB) with a batch size of 16. Adam optimizer is utilized with an initial learning rate of $7.5 \times 10^{-5}$ and a weight decay of $1 \times 10^{-4}$, while the beta values are 0.9 and 0.999 as default.

\subsection{Dataset Details}\label{datasets}
\begin{figure}
    \centering
    \includegraphics[width=0.9\linewidth]{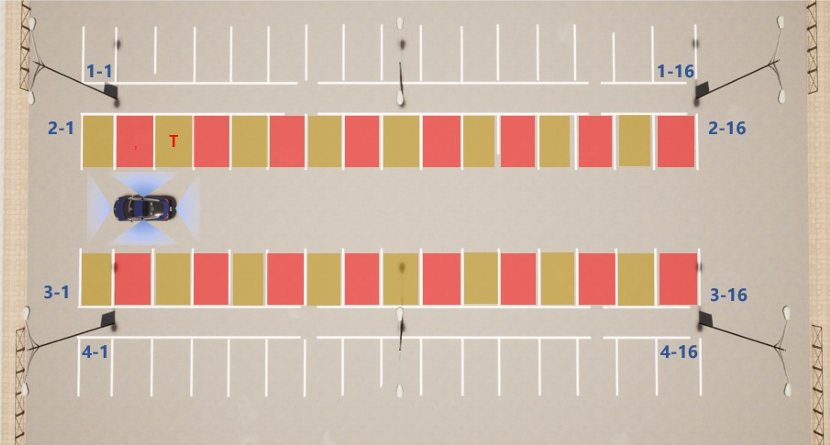}
        \caption{Top view of the parking lot in CARLA.}
    \label{fig:parking_lot}
\end{figure}

\begin{figure*}[htbp]
    \centering
    \subfloat[]{\includegraphics[width=0.24\linewidth]{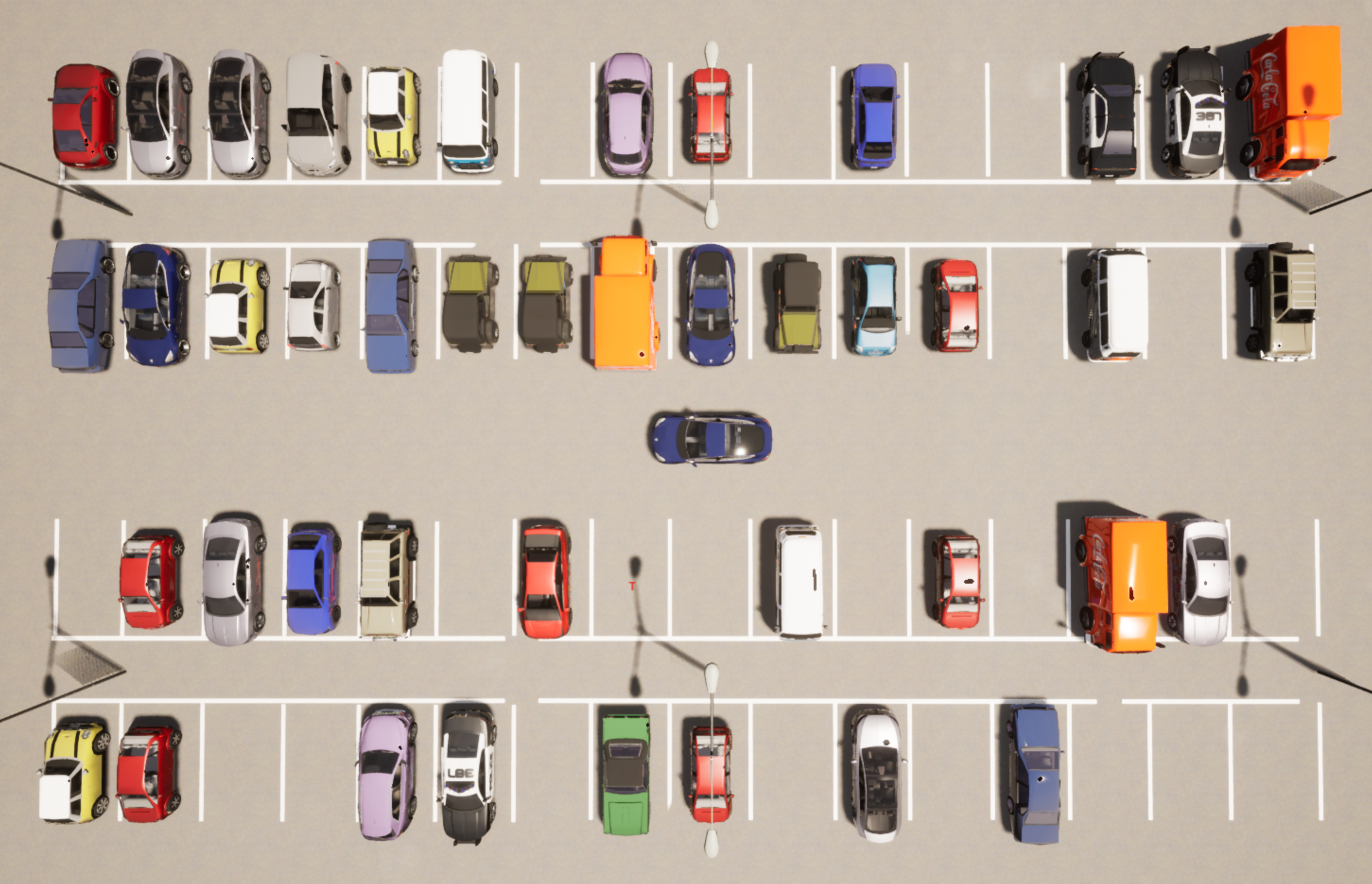}
    \label{fig:seed11_slot3-8_shadow}}
    \subfloat[]{\includegraphics[width=0.24\linewidth]{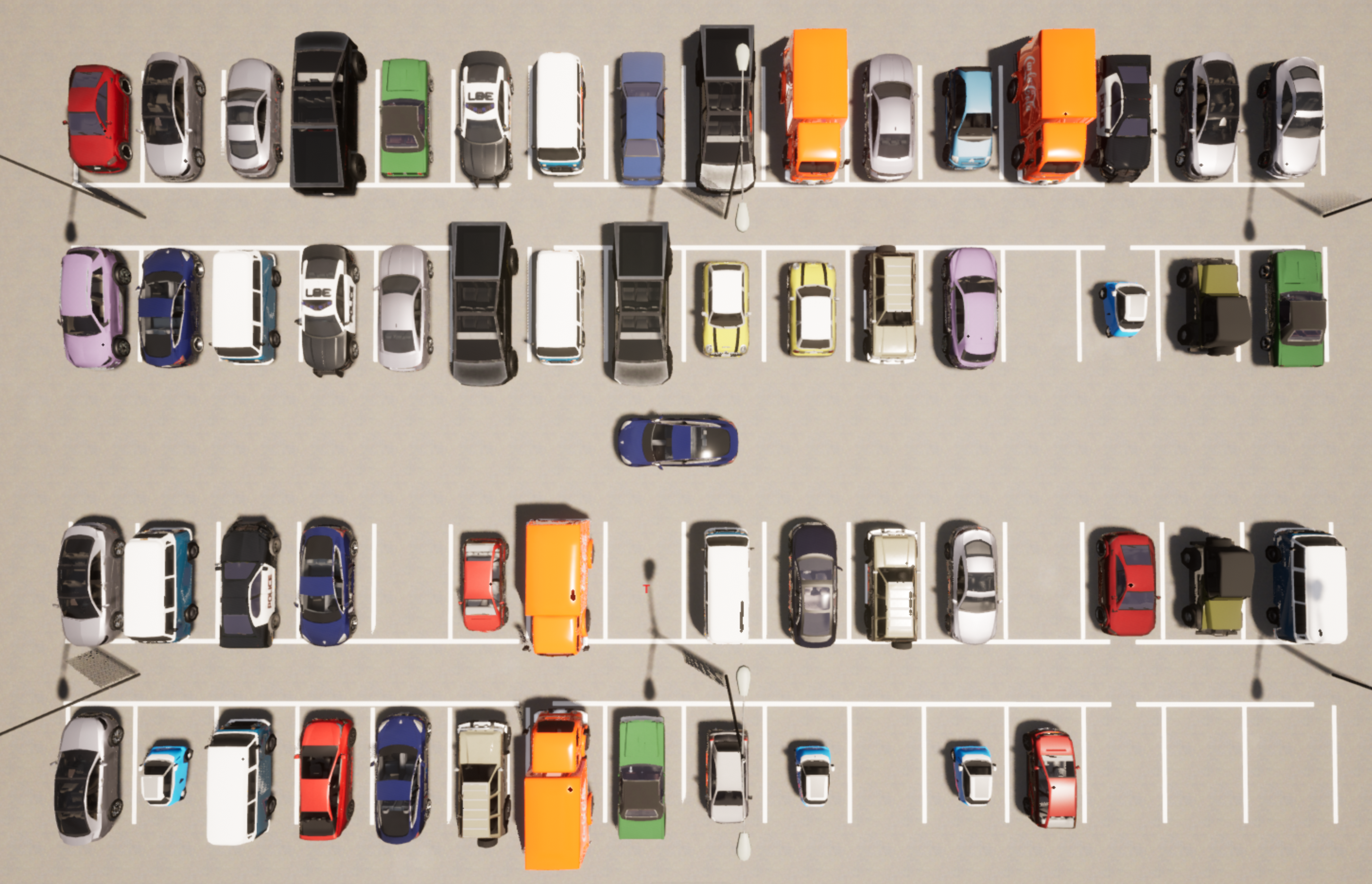}
    \label{fig:seed27_slot3-8_shadow}}
    \subfloat[]{\includegraphics[width=0.24\linewidth]{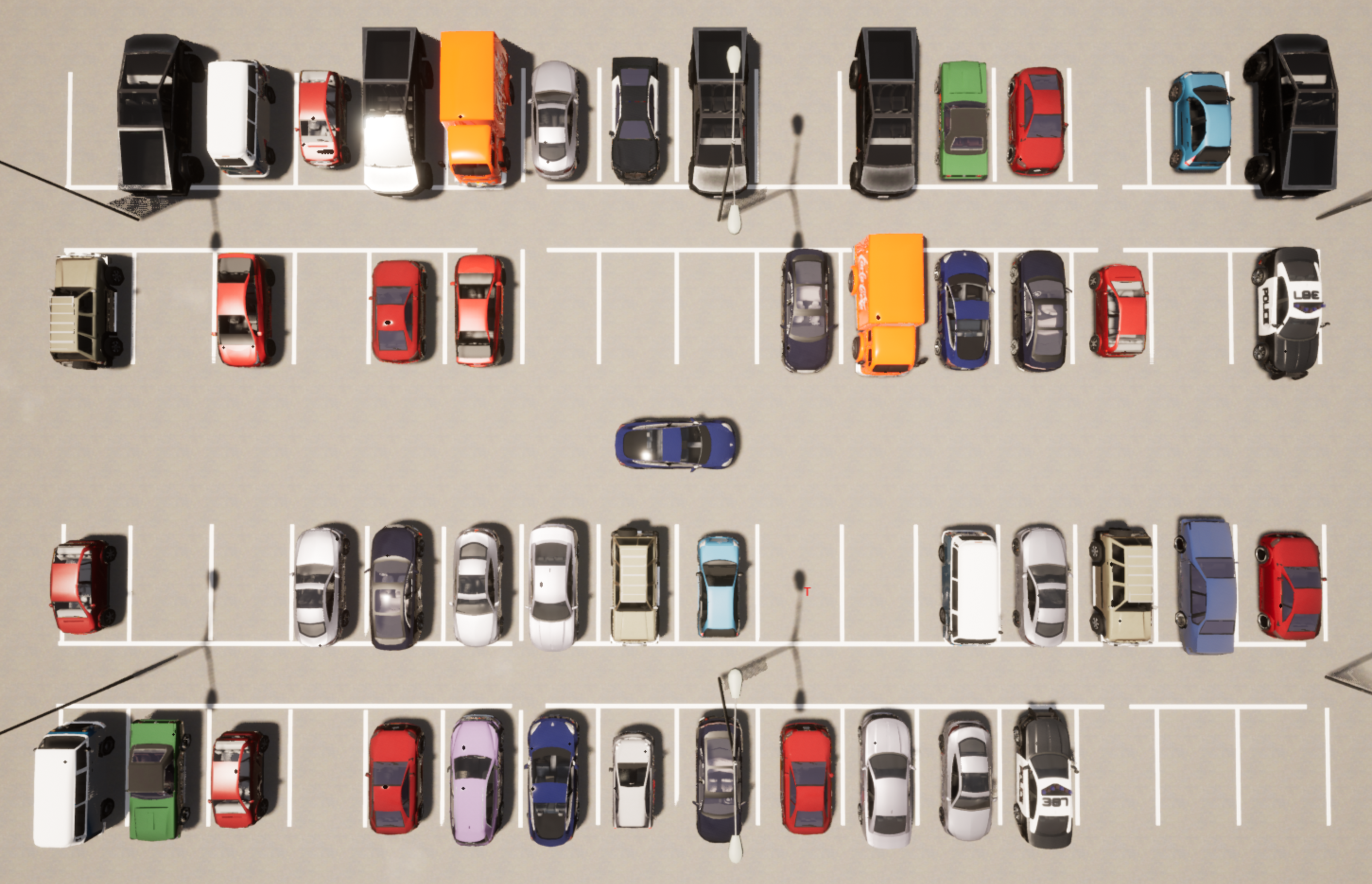}
    \label{fig:seed12_slot3-10_shadow}}
    \subfloat[]{\includegraphics[width=0.24\linewidth]{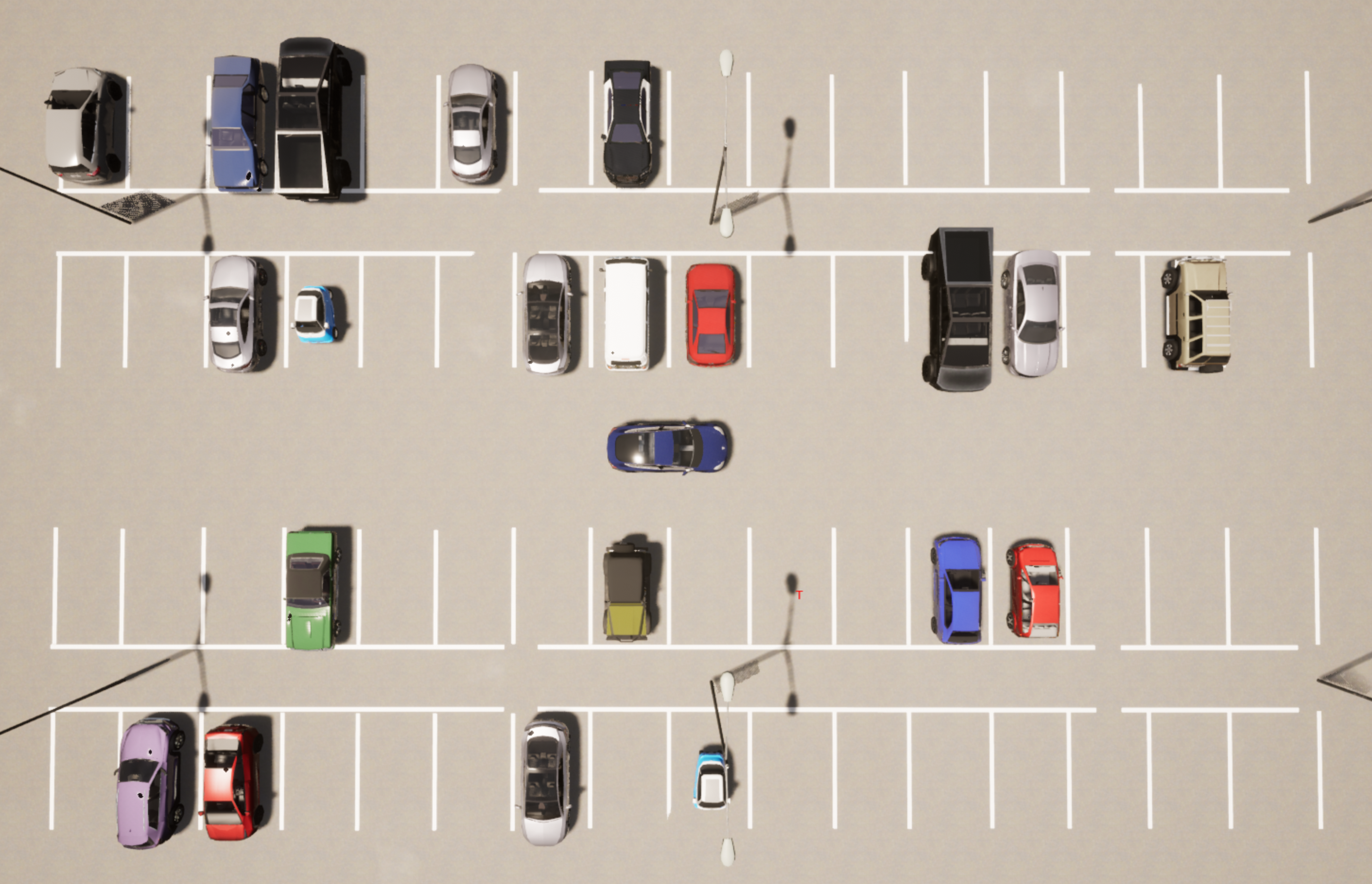}
    \label{fig:seed28_slot3-10_shadow}}
    \caption{
    (a) Scene with random seed = 11, when target slot is 3-8 with shadow. (b) Scene with random seed = 27, when target slot is 3-8 with shadow. (c) Scene with random seed = 12, when target slot is 3-10 with shadow. (d) Scene with random seed = 28, when target slot is 3-10 with shadow.
    }
    \label{fig:scene_examples}
\end{figure*}

The training dataset, Gen 2B of the E2E Parking Dataset~\cite{gao2025e2eparkingdatasetopen}, comprises 248 training and 104 validation trajectories. As illustrated in Fig.~\ref{fig:parking_lot}, training/validation slots (marked in red and brown) utilize even-indexed slots (2-2, \dots, 2-16; 3-2, \dots, 3-16) for expert data generation, while odd-indexed slots (2-1, \dots, 2-15; 3-1, \dots, 3-15) are reserved for closed-loop evaluation. The data generation process is consolidated as follows:

\begin{itemize}
    \item \textbf{Nominal Trajectory Collection (Seeds 0--47):} For each of the 16 slots, scenarios are iteratively instantiated across 48 random seeds. Seeds 0--31 yield 192 training and 64 validation trajectories by querying six fixed BEV initial poses (far, middle, and near across left/right sides) and two distance-randomized lateral poses. Seeds 32--47 contribute 32 training trajectories (far-left/right) and 32 validation trajectories (middle-left/right).
    \item \textbf{Illumination-Adverse Scenarios (Seeds 11, 12, 27, 28):} To introduce visual corner cases, the solar altitude and azimuth angle are dynamically modulated to cast explicit streetlight shadows onto specific slots. Under seeds 11 and 27, shadows are cast on slot 3-8 (Fig.~\ref{fig:seed11_slot3-8_shadow}, \ref{fig:seed27_slot3-8_shadow}); under seeds 12 and 28, shadows are projected onto slot 3-10 (Fig.~\ref{fig:seed12_slot3-10_shadow}, \ref{fig:seed28_slot3-10_shadow}). Each of these four distinct adversarial scenarios contributes 6 training and 2 validation trajectories using the baseline spatial initializations, yielding 24 training and 8 validation trajectories in total.
\end{itemize}

\subsection{Metrics}\label{Metrics}

The metrics used for model testing remain the same as the original ones~\cite{yang2024e2e}.

\textbf{Target Success Rate (TSR)} indicates the probability of the ego vehicle successfully parking in the set target slot. A parking attempt is judged as successful if the vehicle's center is within 0.6 meters horizontally and 1 meter longitudinally from the slot's center, with an oriental error of not higher than 10 degrees.

\textbf{Target Failure Rate (TFR)} indicates the likelihood that the ego vehicle arrives at the designated parking slot but errors beyond the acceptable threshold.

\textbf{Non-Target Rate (NTR)} means the probability that the ego vehicle parks in a slot other than the intended target. \textbf{NTR} is the sum of the \textbf{Non-Target Success Rate (NTSR)} and \textbf{Non-Target Failure Rate (NTFR)}, depending on the final position within the non-target slot meets acceptable error margins.

\textbf{Collision Rate (CR)} measures the frequency of collisions that occur during the parking maneuver.

\textbf{Outbound Rate (OR)} refers to the probability that the ego vehicle exits from the parking lot.

\textbf{Timeout Rate (TR)} indicates the probability that the ego vehicle fails to park successfully within a predefined time constraint (\SI{30}{s}). In the original work, the authors aggregate \textbf{OR} and \textbf{TR}, referring to their sum as \textbf{TR}.

\textbf{Average Position Error (APE)} is the mean Euclidean error between the ego vehicle's final position and the center of the target parking slot in successful parking tasks.

\textbf{Average Orientation Error (AOE)} reflects the average yaw angle deviation between the ego vehicle's final heading and the desired orientation of the parking slot for successful attempts.

\textbf{Average Parking Time (APT)} indicates the mean duration taken for successful parking maneuvers.

\textbf{Average Inference Time (AIT)} represents the average computational time spent per step during model inference.

{
\subsection{Closed-Loop Evaluation Results}\label{Results}

We evaluate the retrained nominal policy and its PSS-supervised version in closed loop in CARLA: executed controls change the vehicle state and hence subsequent observations. Each reported configuration has $16\times6\times4=384$ attempts: 16 held-out target slots, six initial-pose trials per slot, and four evaluation rounds for each slot--pose combination. Thus 384 is the number of attempts, not the number of independent parking layouts or vehicle configurations. Obstacle placement can vary with the scenario seed, but all attempts use the same parking lot, vehicle, sensor arrangement, and success rule. The vehicle has a $30\ \text{s}$ time limit. The tables aggregate attempts across rounds; they do not report per-round variability, so we treat the differences as descriptive results for this test bed rather than evidence of a population-level reliability or cross-environment generalization claim.
}

\begin{figure}[htbp]
    \centering
    \includegraphics[width=\linewidth]{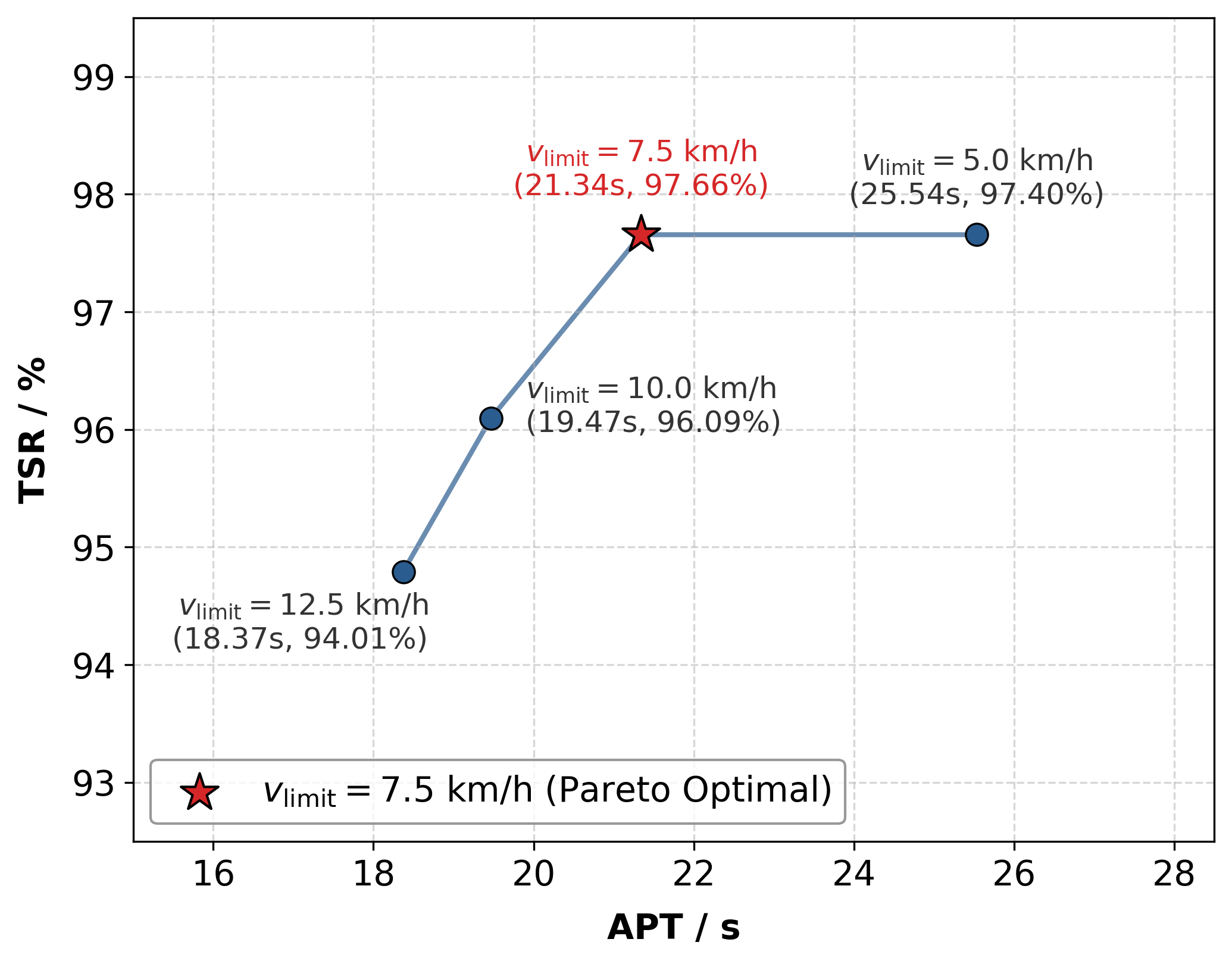}
    \caption{Pareto frontier showcasing the trade-off between TSR and APT under different physical speed limits $v_{\text{limit}}$.}
    \label{fig:pareto_frontier}
\end{figure}

{
\subsubsection{Test with Speed Limit and Pareto Trade-Off}
When the target slot is 2-1 or 3-1 with the ego vehicle facing outward, the baseline model frequently exits the operational zone and acts abnormally. Moreover, due to the slot distribution constraints set by the original authors, the model lacks exposure to streetlight-adjacent slots (e.g., 2-9, 3-9), resulting in delayed braking and subsequent collisions. In rare corner cases, an abnormal control chattering occurs where brake and throttle signals alternate repeatedly post-stopping, inducing a terminal timeout. 

We vary $v_{\text{limit}}$ within the PSS in the same evaluation setting. Figure~\ref{fig:pareto_frontier} reports a TSR of $97.66\%$ and APT of \SI{21.34}{\second} at $7.5\ \mathrm{km/h}$. The $5.0\ \mathrm{km/h}$ setting has the same reported TSR but a longer APT (\SI{25.54}{\second}); the $12.5\ \mathrm{km/h}$ setting has a lower reported TSR ($94.01\%$) and shorter APT (\SI{18.37}{\second}). We use $7.5\ \mathrm{km/h}$ for the remaining results. This selection is specific to the tested lot and does not certify a universally safe speed.
}

\subsubsection{Test with Rule-based Correction}

Table~\ref{tab:Gen_2B_rbc} reports a TSR of 97.656\% (375/384), compared with 85.156\% (327/384) for the same retrained policy without the PSS: a descriptive increase of 12.5 percentage points, or 48 target successes. TR decreases from 7.552\% to 0.781\%, while TFR rises from 0 to 0.260\%; thus the supervisor does not improve every failure category. Among successful attempts, APE changes from 0.237 to 0.211~m and AOE from 0.335$^\circ$ to 0.325$^\circ$. APT changes from 22.080 to 21.341~s. The aggregate table alone does not establish statistical significance, and its accuracy/time means condition on successful attempts whose composition differs between configurations.

At the slot level, reported TSR improves or stays unchanged for 14 of 16 slots and decreases for slots 2-7 and 2-15. The gains at slots 2-1, 2-9, 3-1, and 3-9 are consistent with the specific boundary-exit and delayed-braking failures used to design the guards. Some collisions and timeouts remain in Table~\ref{tab:Gen_2B_rbc}.

\begin{table*}[htbp]
\small
\caption{Closed-loop results for the retrained policy with PSS: 384 repeated attempts ($16$ slots $\times$ $6$ initial-pose trials $\times$ $4$ rounds) in one CARLA lot.}\label{tab:Gen_2B_rbc}%
\renewcommand\arraystretch{1.1}
\centering
\begin{tabular}{@{\hspace{0pt}}m{1.5cm}<{\centering}@{\hspace{0pt}}@{\hspace{0pt}}m{1.5cm}<{\centering}@{\hspace{0pt}}@{\hspace{0pt}}m{1.5cm}<{\centering}@{\hspace{0pt}}@{\hspace{0pt}}m{1.5cm}<{\centering}@{\hspace{0pt}}@{\hspace{0pt}}m{1.5cm}<{\centering}@{\hspace{0pt}}@{\hspace{0pt}}m{1.5cm}<{\centering}@{\hspace{0pt}}@{\hspace{0pt}}m{1.5cm}<{\centering}@{\hspace{0pt}}@{\hspace{0pt}}m{1.5cm}<{\centering}@{\hspace{0pt}}@{\hspace{0pt}}m{1.5cm}<{\centering}@{\hspace{0pt}}@{\hspace{0pt}}m{1.5cm}<{\centering}@{\hspace{0pt}}@{\hspace{0pt}}m{1.5cm}<{\centering}@{\hspace{0pt}}@{\hspace{0pt}}m{1.5cm}<{\centering}@{\hspace{0pt}}}
\toprule
TaskIdx & TSR (\%) & TFR (\%) & NTSR (\%) & NTFR (\%) & CR (\%) & OR (\%) & TR (\%) & APE (m) & AOE (deg) & APT (s) & AIT (s) \\
\midrule
2-1   & 87.500  & 4.167  & 4.167  & 0.000  & 4.167  & 0.000  & 0.000  & 0.199 & 0.600 & 21.156 & 0.077 \\
2-3   & 95.833  & 0.000  & 0.000  & 0.000  & 0.000  & 0.000  & 4.167  & 0.158 & 0.232 & 20.126 & 0.077 \\
2-5   & 100.000 & 0.000  & 0.000  & 0.000  & 0.000  & 0.000  & 0.000  & 0.232 & 0.206 & 20.357 & 0.077 \\
2-7   & 95.833  & 0.000  & 0.000  & 0.000  & 4.167  & 0.000  & 0.000  & 0.247 & 0.204 & 21.181 & 0.077 \\
2-9   & 100.000 & 0.000  & 0.000  & 0.000  & 0.000  & 0.000  & 0.000  & 0.219 & 0.383 & 22.443 & 0.077 \\
2-11  & 100.000 & 0.000  & 0.000  & 0.000  & 0.000  & 0.000  & 0.000  & 0.235 & 0.416 & 20.686 & 0.078 \\
2-13  & 100.000 & 0.000  & 0.000  & 0.000  & 0.000  & 0.000  & 0.000  & 0.196 & 0.308 & 20.896 & 0.078 \\
2-15  & 91.667  & 0.000  & 0.000  & 0.000  & 4.167  & 0.000  & 4.167  & 0.185 & 0.351 & 20.347 & 0.077 \\
3-1   & 100.000 & 0.000  & 0.000  & 0.000  & 0.000  & 0.000  & 0.000  & 0.233 & 0.668 & 21.694 & 0.077 \\
3-3   & 100.000 & 0.000  & 0.000  & 0.000  & 0.000  & 0.000  & 0.000  & 0.170 & 0.321 & 21.574 & 0.078 \\
3-5   & 100.000 & 0.000  & 0.000  & 0.000  & 0.000  & 0.000  & 0.000  & 0.267 & 0.482 & 22.136 & 0.078 \\
3-7   & 100.000 & 0.000  & 0.000  & 0.000  & 0.000  & 0.000  & 0.000  & 0.169 & 0.132 & 20.804 & 0.078 \\
3-9   & 100.000 & 0.000  & 0.000  & 0.000  & 0.000  & 0.000  & 0.000  & 0.270 & 0.248 & 20.517 & 0.078 \\
3-11  & 100.000 & 0.000  & 0.000  & 0.000  & 0.000  & 0.000  & 0.000  & 0.208 & 0.231 & 21.758 & 0.077 \\
3-13  & 95.833  & 0.000  & 0.000  & 0.000  & 0.000  & 0.000  & 4.167  & 0.139 & 0.200 & 23.896 & 0.077 \\
3-15  & 95.833  & 0.000  & 0.000  & 0.000  & 4.167  & 0.000  & 0.000  & 0.243 & 0.221 & 21.883 & 0.077 \\
\textbf{Avg} & \textbf{97.656} & \textbf{0.260} & \textbf{0.260} & \textbf{0.000} & \textbf{1.042} & \textbf{0.000} & \textbf{0.781} & \textbf{0.211} & \textbf{0.325} & \textbf{21.341} & \textbf{0.077} \\
\bottomrule
\end{tabular}
\end{table*}

\subsubsection{Visualization of Comparisons}

\paragraph{Parking Slot 2-1}

\begin{figure}[htbp]
    \centering
    \subfloat[Case a]{\includegraphics[width=0.4\linewidth]{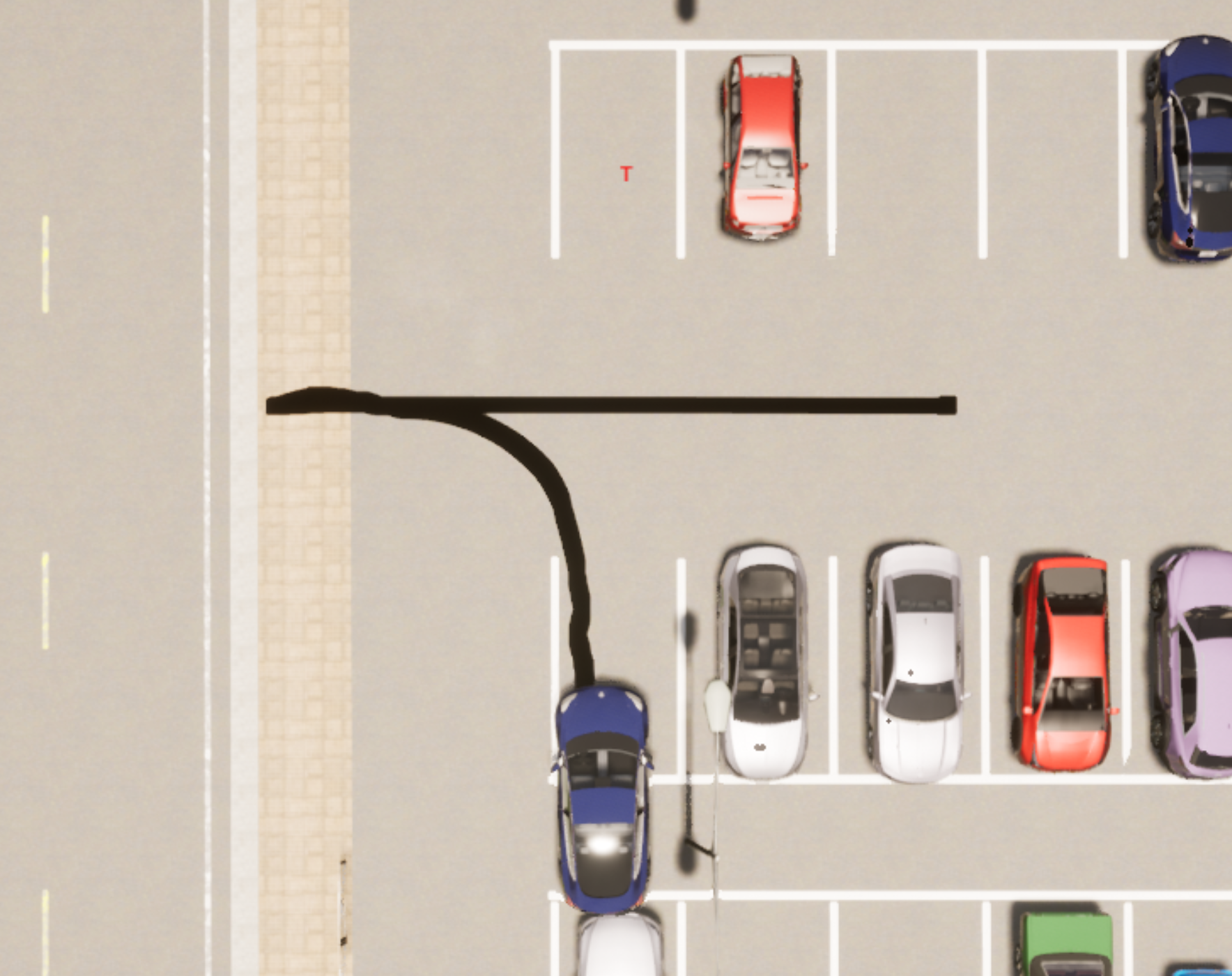} \label{fig:accident1}}
    \subfloat[Case b]{\includegraphics[width=0.4\linewidth]{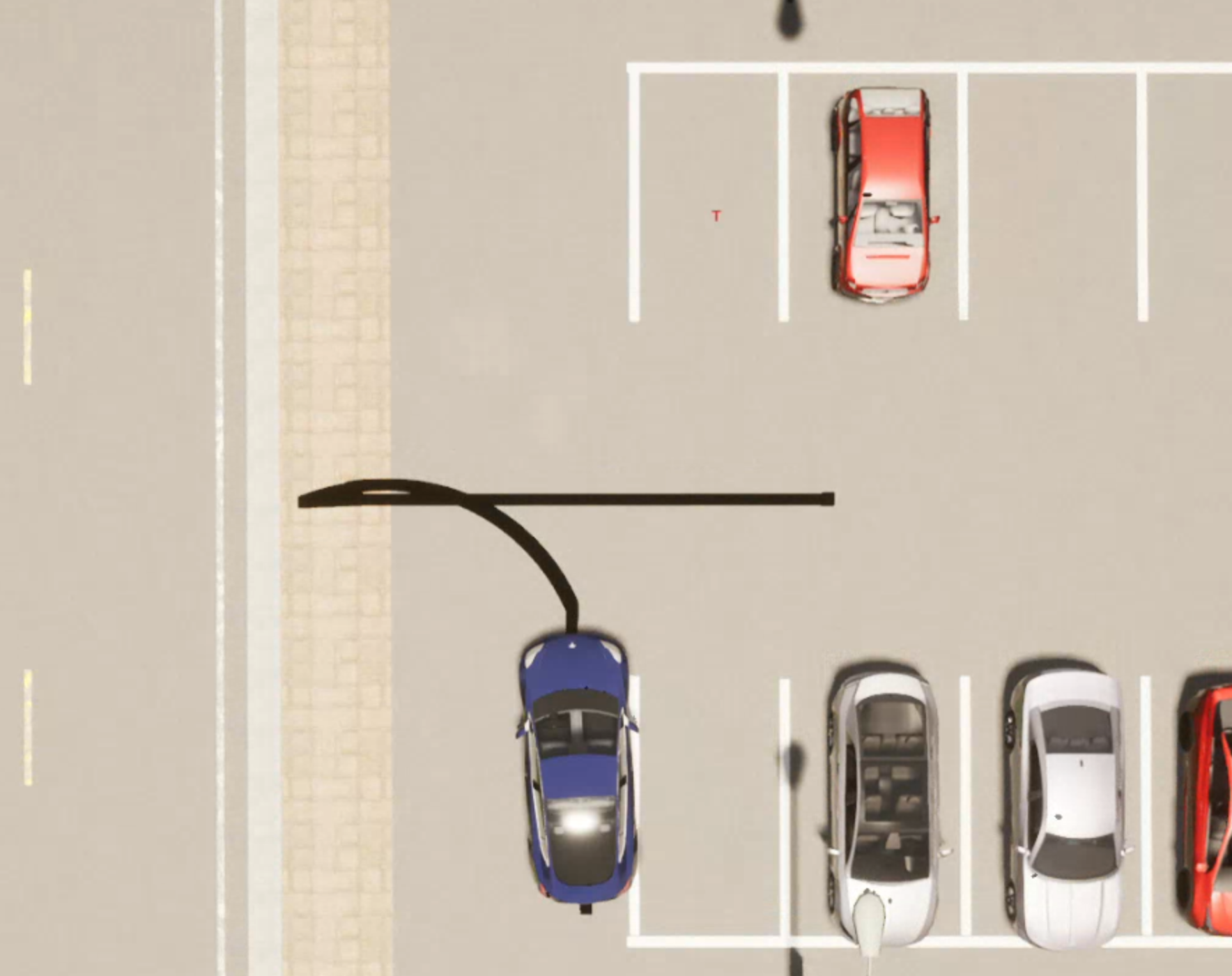} \label{fig:accident2}}\\
    \subfloat[Case c]{\includegraphics[width=0.4\linewidth]{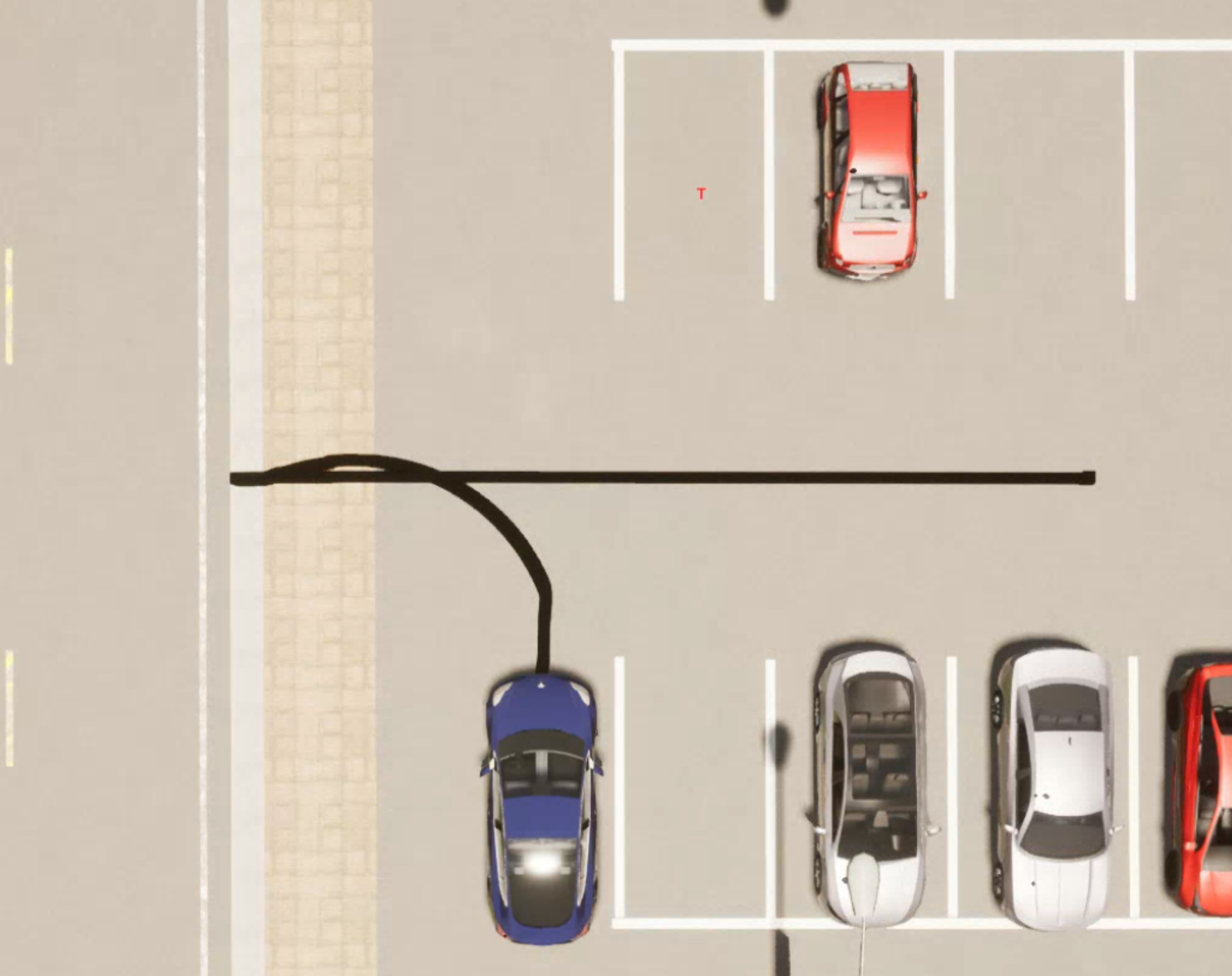} \label{fig:accident3}}
    \subfloat[Case d]{\includegraphics[width=0.4\linewidth]{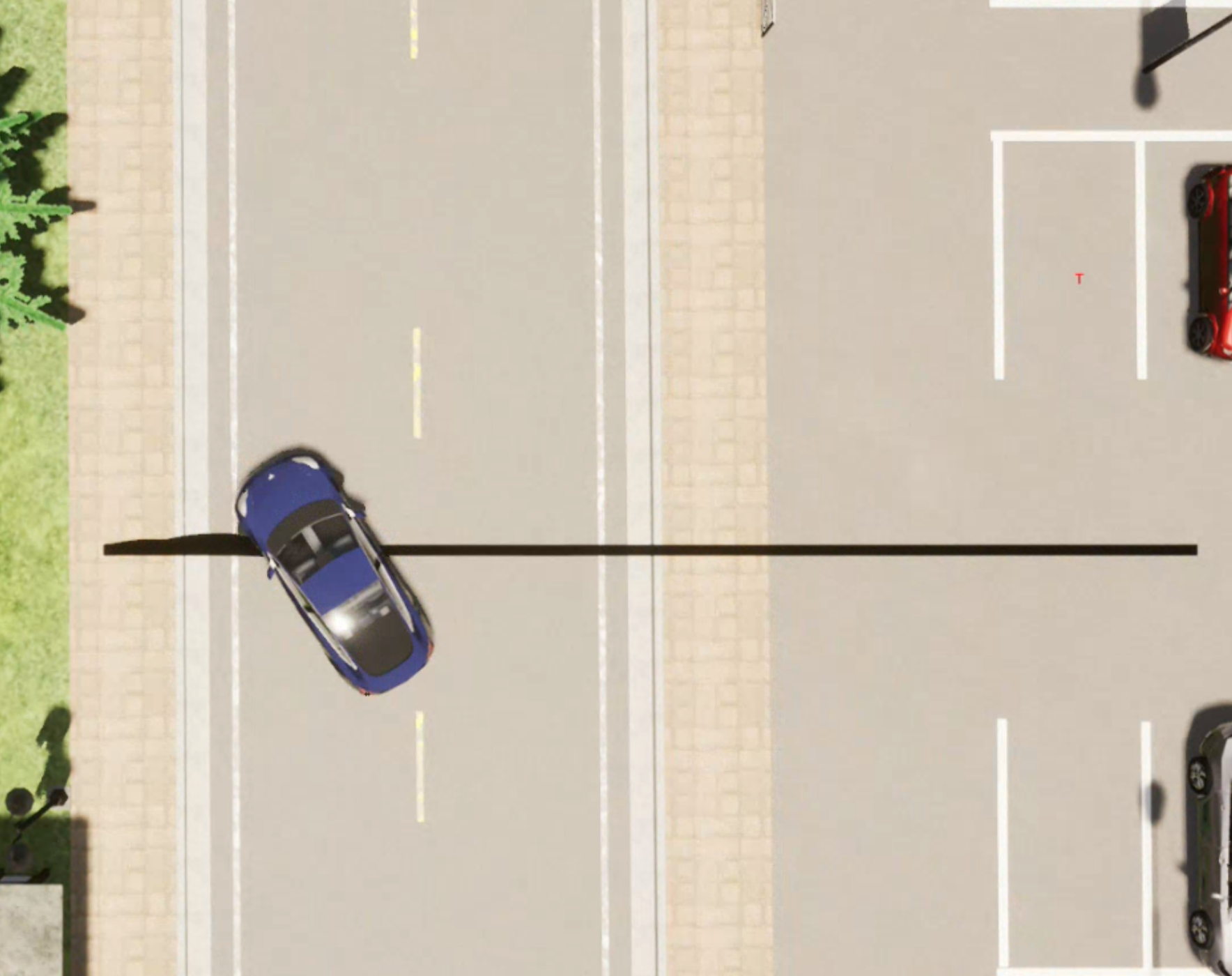} \label{fig:accident4}}
    \caption{Visualization of four failure cases.}
    \label{fig:accident_all}
\end{figure}

Figure~\ref{fig:accident_all} illustrate multiple failure cases during the closed-loop test with the original model, when the target is parking slot 2-1. The model has various incorrect inferences when the vehicle is located at the edge of the parking lot, where a typical problem is that it cannot track the target slot. In Figure~\ref{fig:accident1}, the vehicle fails to stop in a non-target slot 3-1 with activated throttle and collides with the static vehicle at the back. In Figure~\ref{fig:accident2} and Figure~\ref{fig:accident3}, the vehicle stops with activated brake outside of a non-target slot 3-1, as if there were an additional slot between slot 3-1 and the sidewalk. In Figure~\ref{fig:accident4}, the vehicle exits from the parking lot and it attempts to be parked on the road, which is regarded as an extremely dangerous operation.

\begin{figure}
    \centering
    \includegraphics[width=0.4\linewidth]{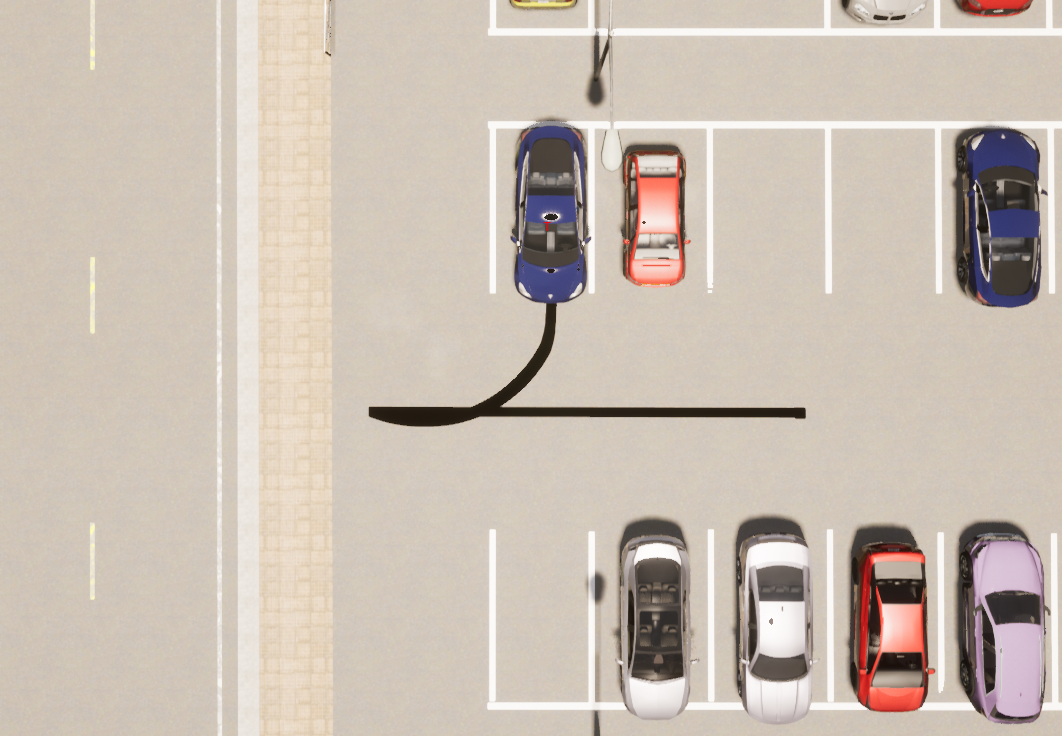}
    \caption{Visualization of a corrected trajectory.}
    \label{fig:corrected_trajectory}
\end{figure}

Figure~\ref{fig:corrected_trajectory} depicts the corrected trajectory with the help of rule-based correction. With an earlier braking point, the vehicle stops at the correct location, and the subsequent 2.5-second timed automaton signal output is sufficient to correct its behavior, enabling it to complete the following parking maneuver without any further intervention. From the above cases, it can be judged that rule-based correction is necessary to get deployed to constrain the control signals from model inference.

\begin{figure}[h]
    \centering
    \subfloat[$T = \SI{4.5}{\second}$]{\includegraphics[width=0.25\linewidth]{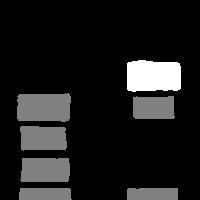} \label{fig:timeout_0135}}
    \hfill
    \subfloat[$T = \SI{4.6}{\second}$]{\includegraphics[width=0.25\linewidth]{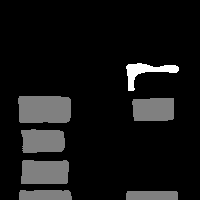} \label{fig:timeout_0138}}
    \hfill
    \subfloat[$T = \SI{5.1}{\second}$]{\includegraphics[width=0.25\linewidth]{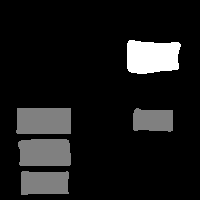} \label{fig:timeout_0153}}\\

    \subfloat[$T = \SI{8.8}{\second}$]{\includegraphics[width=0.25\linewidth]{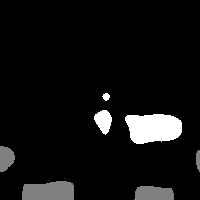} \label{fig:timeout_0264}}
    \hfill
    \subfloat[$T = \SI{9.5}{\second}$]{\includegraphics[width=0.25\linewidth]{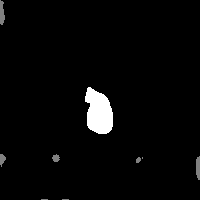} \label{fig:timeout_0285}}
    \hfill
    \subfloat[$T = \SI{16.8}{\second}$]{\includegraphics[width=0.25\linewidth]{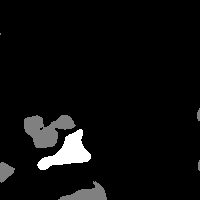} \label{fig:timeout_0504}}\\

    \subfloat[$T = \SI{17.6}{\second}$]{\includegraphics[width=0.25\linewidth]{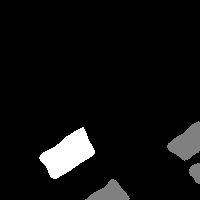} \label{fig:timeout_0528}}
    \hfill
    \subfloat[$T = \SI{21.8}{\second}$]{\includegraphics[width=0.25\linewidth]{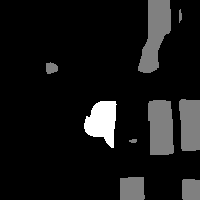} \label{fig:timeout_0654}}
    \hfill
    \subfloat[$T = \SI{30}{\second}$]{\includegraphics[width=0.25\linewidth]{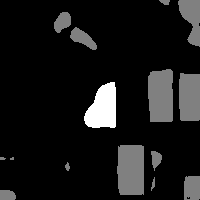} \label{fig:timeout_0900}}
    
    \caption{BEV features of a timeout case.}
    \label{fig:BEV_timeout}
\end{figure}

The failure cases can be explained by the generated BEV features. Figure~\ref{fig:BEV_timeout} displays BEV features of a timeout case in different time steps, where drivable area is black, predicted target slot is write and obstacles are grey. The target in the test is slot 2-1. The BEV feature remains to be correct until $T = \SI{4.5}{\second}$. After that, the segmentation output of the target parking slot disappears and moves to the area between slot 2-1 and sidewalk. At $T = \SI{9.5}{\second}$, the vehicle stops at the edge of the parking lot and the segmentation of target is located at the vehicle's current position. At $T = \SI{16.8}{\second}$ the vehicle starts to reverse in the opposite direction of the target and the BEV feature is confusing. In the end the vehicle stops at the area between slot 3-1 and sidewalk, with the predicted target segmentation at the same position away from the ground-truth target. Meanwhile, the segmentation of other vehicles is problematic, which also leads to incorrect output signals.

\begin{figure}[b]
    \centering
    \subfloat[$T = \SI{4.0}{\second}$]{\includegraphics[width=0.25\linewidth]{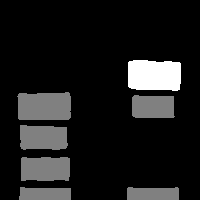} \label{fig:outbound_0120}}
    \hfill
    \subfloat[$T = \SI{4.1}{\second}$]{\includegraphics[width=0.25\linewidth]{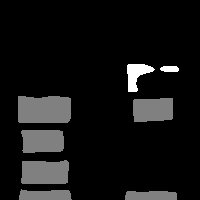} \label{fig:outbound_0123}}
    \hfill
    \subfloat[$T = \SI{4.4}{\second}$]{\includegraphics[width=0.25\linewidth]{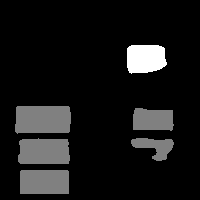} \label{fig:outbound_0132}}\\

    \subfloat[$T = \SI{6.8}{\second}$]{\includegraphics[width=0.25\linewidth]{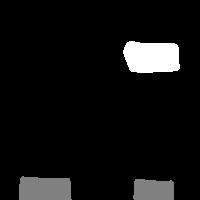} \label{fig:outbound_0204}}
    \hfill
    \subfloat[$T = \SI{8.6}{\second}$]{\includegraphics[width=0.25\linewidth]{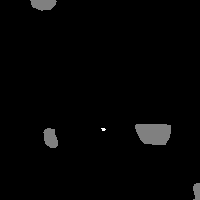} \label{fig:outbound_0258}}
    \hfill
    \subfloat[$T = \SI{9.5}{\second}$]{\includegraphics[width=0.25\linewidth]{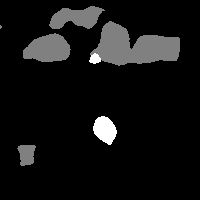} \label{fig:outbound_0285}}
    
    \caption{BEV features of an outbound case.}
    \label{fig:BEV_outbound}
\end{figure}

Figure~\ref{fig:BEV_outbound} demonstrates BEV features of a outbound case in different time steps. The BEV features in the first 5 seconds remain the same as those in the timeout case. At $T = \SI{8.6}{\second}$ the target segmentation moves outside of the parking lot and the vehicle continues moving out of the parking lot without slowing down. Afterwards, the vehicle approaches the green area and target segmentation is abnormal. It disappears unexpectedly and then appears at the vehicle's current position, which is far away from the ground-truth target.

\begin{figure}
    \centering
    \includegraphics[width=0.9\linewidth]{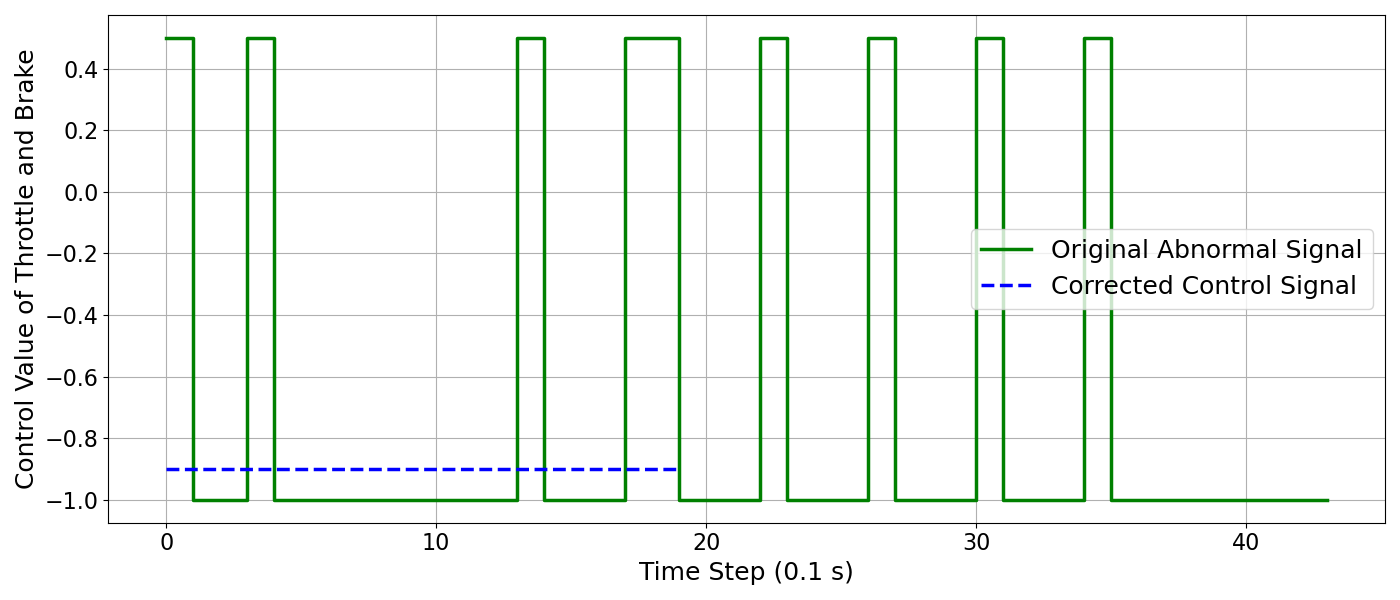}
    \caption{Comparison of abnormal and corrected control signal.}
    \label{fig:control_value}
\end{figure}

\begin{figure}[htbp]
    \centering

    \subfloat[$T = \SI{8.6}{\second}$]{\includegraphics[width=0.25\linewidth]{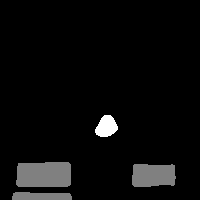} \label{fig:rule_0258}}
    \hfill
    \subfloat[$T = \SI{9.1}{\second}$]{\includegraphics[width=0.25\linewidth]{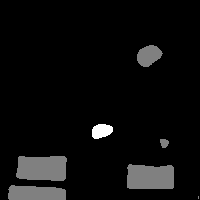} \label{fig:rule_0273}}
    \hfill
    \subfloat[$T = \SI{9.4}{\second}$]{\includegraphics[width=0.25\linewidth]{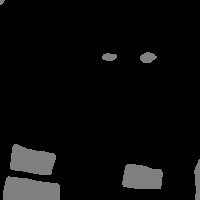} \label{fig:rule_0282}}

    \subfloat[$T = \SI{9.5}{\second}$]{\includegraphics[width=0.25\linewidth]{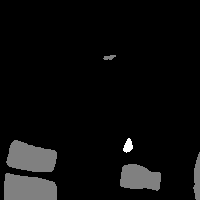} \label{fig:rule_0285}}
    \hfill
    \subfloat[$T = \SI{9.8}{\second}$]{\includegraphics[width=0.25\linewidth]{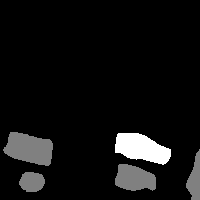} \label{fig:rule_0294}}
    \hfill
    \subfloat[$T = \SI{14.0}{\second}$]{\includegraphics[width=0.25\linewidth]{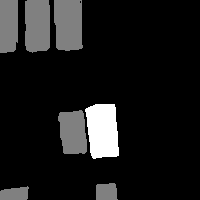} \label{fig:rule_0420}}
    
    \caption{BEV features corrected by rules.}
    \label{fig:BEV_rule}
\end{figure}

Figure~\ref{fig:BEV_rule} demonstrates BEV features corrected by rules. The transformation of BEV feature at an early stage is similar to the above 2 cases shown in Figure~\ref{fig:timeout_0135}, \ref{fig:timeout_0138} and \ref{fig:timeout_0153} or Figure~\ref{fig:outbound_0120}, \ref{fig:outbound_0123} and \ref{fig:outbound_0132}. When the vehicle stops at the edge of the parking lot, the target segmentation disappears and the target becomes unknown to the model. In the process of forced reverse set by rule, the target slot area can be gradually segmented, which leads to a success parking in the end. The segmentation of target becomes explicit and correct after the activation of timed automaton.

\paragraph{Parking Slot 3-9}

Figure~\ref{fig:control_value} compares the control signals with and without the intervention of rule-based correction after stopping of the vehicle. The y-axis is control value of throttle and brake, where positive values stand for throttle and negative values indicate brake. Although the vehicle is parked into the target slot with allowable positional and oriental errors, the vehicle may frequently get throttle signals after it stops, which is certainly considered as a dangerous behavior. This alternating output of throttle and brake is judged as a timeout, since maintaining a throttle output of zero for 2 seconds is a necessary condition for determining a successful parking. With the help of rule-based correction, the brake value remains to be 0.9 in order to force the vehicle to keep static state.

\begin{figure}[htbp]
    \centering

    \subfloat[$T = \SI{16.8}{\second}$]{\includegraphics[width=0.25\linewidth]{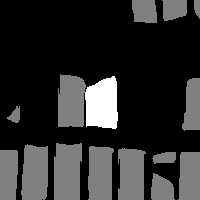} \label{fig:collision_0504}}
    \hfill
    \subfloat[$T = \SI{17.3}{\second}$]{\includegraphics[width=0.25\linewidth]{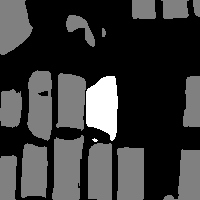} \label{fig:collision_0519}}
    \hfill
    \subfloat[$T = \SI{18.2}{\second}$]{\includegraphics[width=0.25\linewidth]{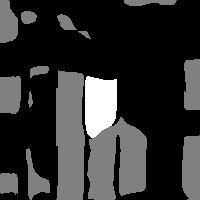} \label{fig:collision_0546}}    
    \caption{BEV features before collision at target slot 3-9.}
    \label{fig:BEV_collision}
\end{figure}

\begin{figure}[htbp]
    \centering

    \subfloat[$T = \SI{15.8}{\second}$]{\includegraphics[width=0.25\linewidth]{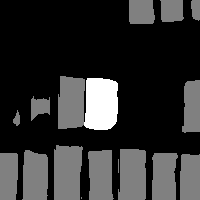} \label{fig:brake_0474}}
    \hfill
    \subfloat[$T = \SI{17.3}{\second}$]{\includegraphics[width=0.25\linewidth]{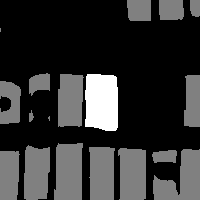} \label{fig:brake_0519}}
    \hfill
    \subfloat[$T = \SI{18.7}{\second}$]{\includegraphics[width=0.25\linewidth]{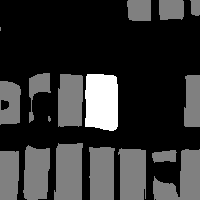} \label{fig:brake_0561}}    
    \caption{BEV features during forced brake at target slot 3-9.}
    \label{fig:BEV_brake}
\end{figure}

\begin{figure}[htbp]
     \centering

    \subfloat[$T = \SI{10.7}{\second}$]{\includegraphics[width=0.25\linewidth]{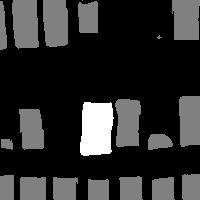} \label{fig:abnormal_0321}}
    \hfill
    \subfloat[$T = \SI{14.6}{\second}$]{\includegraphics[width=0.25\linewidth]{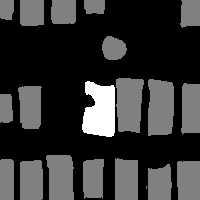} \label{fig:abnormal_0438}}
    \hfill
    \subfloat[$T = \SI{18.3}{\second}$]{\includegraphics[width=0.25\linewidth]{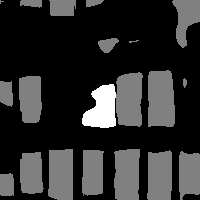} \label{fig:abnormal_0549}}   
    \caption{Abnormal BEV features during forced brake at target slot 3-9.}
    \label{fig:BEV_abnormal_brake}
\end{figure}

The BEV feature may become abnormal in the end of the parking task. Since there is no segmentation label for streetlight, it is challenging to avoid collision with streetlight. In Figure~\ref{fig:BEV_collision}, the segmentation result is distorted and the vehicle fails to stop at the right moment, colliding with the post of a streetlight. In Figure~\ref{fig:BEV_brake}, the vehicle is forced to get braked at the designated position and the segmentation remains to be normal and explicit. In some cases where BEV features are abnormal (like Figure~\ref{fig:BEV_abnormal_brake}), the PSS overwrites the model's wrong control signals influenced by distorted BEV features and forces the vehicle to stop at the right position. If there were no PSS, the shrinking of the target segmentation may cause unexpected problems.

\subsubsection{Comparison Scope and Recent Parking Methods}

Table~\ref{comparison_baseline_expert} retains the reported Yang et al. model and expert results for context. Only the two Gen 2B policy rows isolate the PSS within the evaluation reported here. The Yang et al. models, expert demonstrations, and our policy use different data/model availability or evaluation conditions; numerical differences between those rows cannot be attributed to the PSS. The expert rows are demonstration quality references, not deployable policy baselines.

Table~\ref{tab:recent_methods} compares our approach with two recent peer-reviewed parking methods at the level of method, output, and evaluation domain. ParkingE2E~\cite{li2024parkinge2e} learns image-to-waypoint planning and reports 87.8\% mean success across four real garages. HOPE~\cite{jiang2025hope} learns a hybrid path planner using reinforcement learning and Reeds--Shepp curves, tests difficulty-stratified scenarios, and includes a real-world demonstration. Our method instead modifies low-level controls from an inherited network and is evaluated in one CARLA lot. Different sensors, outputs, vehicles, success definitions, and environments prevent a defensible ranking from their published rates. Reimplementing both methods and evaluating them under a shared protocol would be required to satisfy a direct head-to-head comparison; such experiments are not part of the available results.

\begin{table*}[t]
\caption{Methodological comparison with recent peer-reviewed parking work. Published evaluations use different protocols; this table is not a head-to-head performance comparison.}
\label{tab:recent_methods}
\centering\footnotesize
\renewcommand{\arraystretch}{1.35}
\begin{tabularx}{\textwidth}{p{25mm}p{35mm}p{38mm}X}
\toprule
\textbf{Method} & \textbf{Core decision mechanism} & \textbf{Output and inputs} & \textbf{Reported evaluation scope} \\
\midrule
ParkingE2E~\cite{li2024parkinge2e} & Learned target-query Transformer & Future waypoints from surround-view cameras & Four real garages and real-vehicle tests; 87.8\% mean parking success under its own protocol \\
\addlinespace
HOPE~\cite{jiang2025hope} & Reinforcement learning plus Reeds--Shepp curves and action masking & Parking path from its planner state representation & Difficulty-stratified parking scenarios and real-world experiments \\
\addlinespace
This work & Inherited E2E Parking Transformer plus hand-coded timed output supervisor & Low-level controls from cameras, target, vehicle state, and CARLA world coordinates & One CARLA lot, one vehicle/sensor setup; 16 slots $\times$ 6 initial poses $\times$ 4 rounds \\
\bottomrule
\end{tabularx}
\end{table*}

\begin{table*}[htbp]
\small
\caption{Reported policy and expert results for context. Only the two Gen 2B policy rows are the within-study comparison of the PSS.}\label{comparison_baseline_expert}%
\renewcommand\arraystretch{1.1}
\centering
\begin{tabular}{@{\hspace{0pt}}m{6cm}<{\centering}@{\hspace{0pt}}m{2cm}<{\centering}@{\hspace{0pt}}m{2cm}<{\centering}@{\hspace{0pt}}m{2cm}<{\centering}@{\hspace{0pt}}m{2cm}<{\centering}@{\hspace{0pt}}m{2cm}<{\centering}@{\hspace{0pt}}m{2cm}<{\centering}@{\hspace{0pt}}}
\toprule
TaskIdx & TSR (\%)  & TFR (\%)  & CR (\%)  & APE (m)  & AOE (deg) & APT (s) \\
\midrule
E2E Parking~\cite{yang2024e2e} Unreleased Model & 91.41 & 2.08 & 2.08 & 0.30 & 0.87 & 15.72 \\
E2E Parking~\cite{yang2024e2e} Released Model & 75.00 & 1.56 & 10.42 & 0.25 & 0.63 & 18.41 \\
Model trained on Gen 2B~\cite{gao2025e2eparkingdatasetopen} (Baseline)   & 85.16 & 0.00 & 2.87 & 0.24 & 0.34 & 22.08 \\
\textbf{Model trained on Gen 2B~\cite{gao2025e2eparkingdatasetopen} + Rule-based Correction}   & \textbf{97.66} & 0.26 & 1.04 & 0.21 & 0.33 & 21.34 \\
Original Expert~\cite{yang2024e2e}   & 100.00 & 0.00 & 0.00 & 0.23 & 0.48 & 14.96 \\
\textbf{New Expert} &  100.00 & 0.00 & 0.00 & 0.13 & 0.05 & 16.95 \\
\bottomrule
\end{tabular}
\end{table*}

{
\subsubsection{Failure Case Analysis and Limitations}
Closed-loop evaluations reveal two primary limitations of the proposed framework. First, as qualitatively illustrated in Fig.~\ref{fig:timeout}, if the baseline model severely overshoots the designated slot, the PSS successfully prevents out-of-boundary run-off but lacks a proximity-based fall-back mechanism to command an immediate stop, causing the vehicle to eventually park in an incorrect non-target slot. Second, the system occasionally suffers from control chattering in obstacle-free paths, where conflicting neural outputs and conservative PSS regulations trigger alternating throttle and braking cycles. While the vehicle is still guided safely into the target slot, this efficiency degradation ultimately results in a terminal timeout failure.

\begin{figure}
    \centering
    \includegraphics[width=0.5\linewidth]{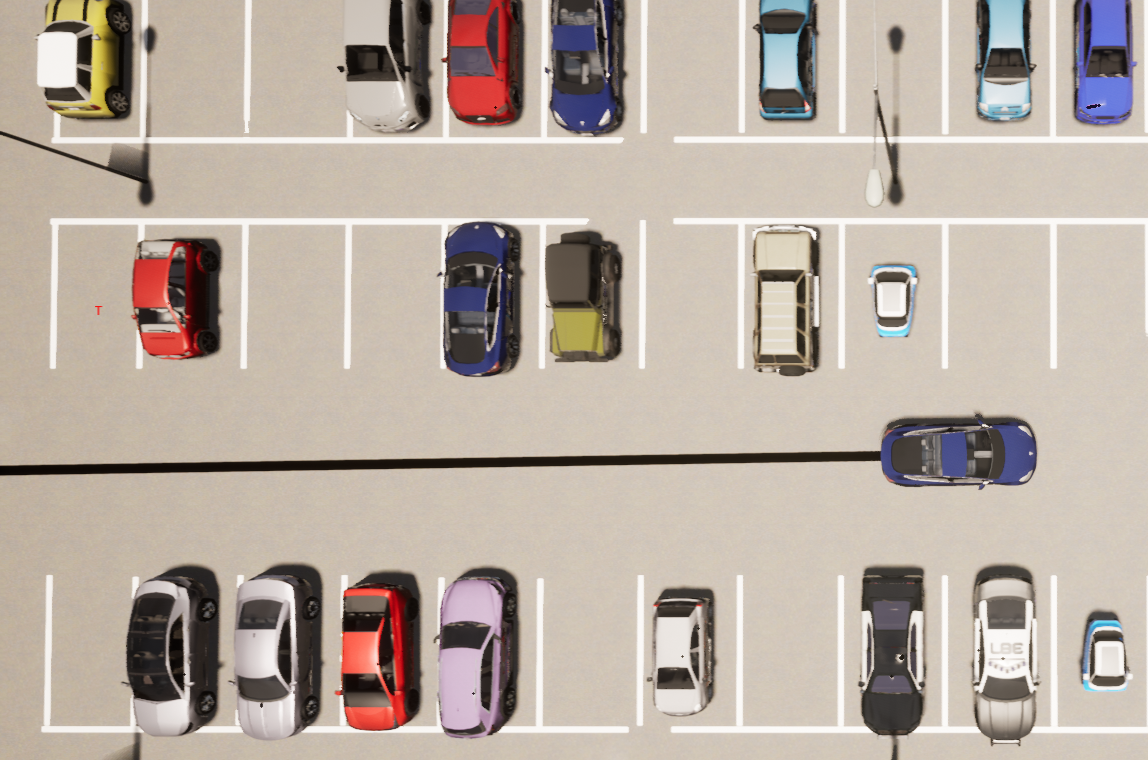}
    \caption{Example of a closed-loop failure after target-slot overshoot.}
    \label{fig:timeout}
\end{figure}

\subsection{Ablation Study}\label{sec:ablation}
We ablate three guard components using the same $16\times6\times4$ evaluation structure. Table~\ref{tab:ablation_results} reports lower TSR after removing target alignment (93.49\%), speed limiting (91.93\%), or geofencing (89.84\%) than with the complete PSS (97.66\%). The respective APTs are 21.66, 18.07, 21.54, and 21.34~s. These descriptive results support the usefulness of the tested guards in this lot. They do not show that each guard is necessary in every parking environment, nor do they prove safety.
}

\begin{table}[htbp]
\caption{PSS component ablations over 384 repeated attempts per configuration in one CARLA lot}
\label{tab:ablation_results}
\centering
\small
\begin{tabular}{lcc}
\toprule
\textbf{Configuration} & \textbf{TSR (\%)} & \textbf{APT (s)} \\
\midrule
(i) w/o Target Alignment  & 93.49 & 21.66  \\
(ii) w/o Speed Limit     & 91.93 & 18.07  \\
(iii) w/o Geofencing      & 89.84 & 21.54  \\
\midrule
\textbf{(iv) Complete PSS} & \textbf{97.66} & 21.34 \\
\bottomrule
\end{tabular}
\end{table}

\section{Conclusion}\label{Conclusion}

This study adds a hand-designed timed output supervisor to the existing E2E Parking network. In our single-lot CARLA evaluation, the retrained policy's target successes increase from 327 to 375 out of 384 repeated attempts with the PSS. The ablations associate the observed improvement with several guards, but neither these results nor the state-machine specification establish a formal safety guarantee or generalization beyond this configuration.

The rules were selected after manual inspection of failures and include fixed actuator values, durations, and map-specific coordinates. All 384 attempts per configuration arise from only 16 slots, six pose trials, and four rounds in the same CARLA lot, vehicle, and sensor setup. Four rounds do not characterize reliability across independently sampled environments; no different lot, vehicle, sensor suite, or real-world test was performed. The PSS uses simulator world coordinates that a real vehicle would need to estimate with uncertainty. Our earlier experiment with target coordinates in the vehicle frame was less stable, and no robust localization replacement is demonstrated here. Finally, a direct same-protocol experiment against ParkingE2E or HOPE remains outstanding.

Future evaluation should report round- and scenario-level outcomes and test more independently sampled lots, obstacle layouts, vehicles, and sensors. A controlled comparison with recent parking methods requires a common scenario set and success definition. Before real-vehicle use, the guards also need a localization source, uncertainty handling, and safety analysis.

To support reproducibility and further research, we release the full implementation of our framework. We hope this work encourages more exploration into the coordination between visual-based end-to-end models and rule-based traditional controllers in autonomous driving, especially in structured, low-speed and low-risk parking scenarios.


 
%

\bibliographystyle{IEEEtran}
\bibliography{sn-bibliography}

\vspace{11pt}

\input{biography}






\end{document}

%% file: biography.tex
\begin{IEEEbiography}[{\includegraphics[width=1in,height=1.25in,clip,keepaspectratio]{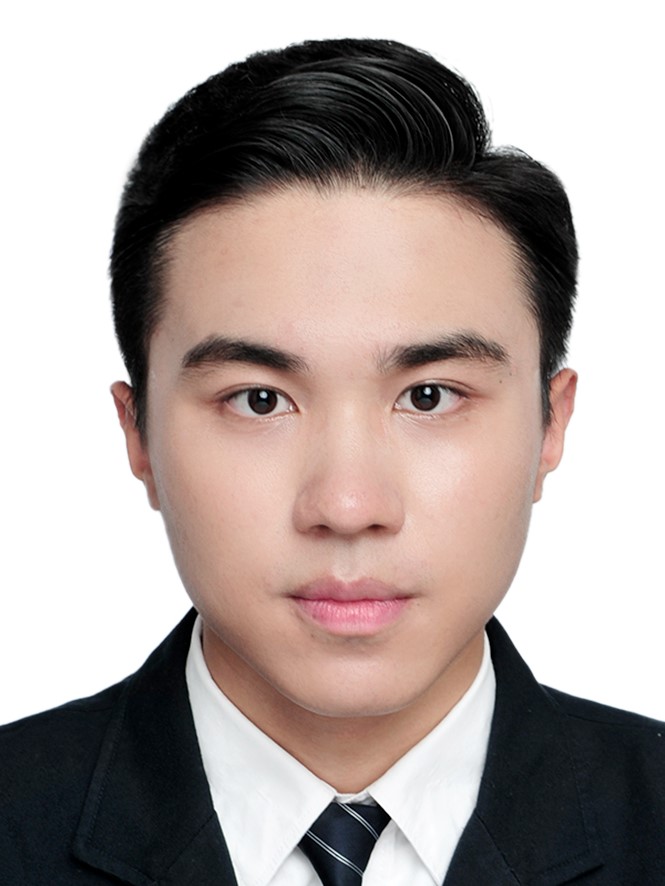}}]{Kejia Gao}
received a bachelor's degree in Mechatronics from Tongji University in 2022 and a master's degree in Robotics, Cognition, and Intelligence from the Technical University of Munich in 2025. He is currently working at the Shanghai Institute of Optics and Fine Mechanics, Chinese Academy of Sciences. His research focuses on computer vision and AI-driven optics.
\end{IEEEbiography}

\vspace{11pt}

\begin{IEEEbiography}[{\includegraphics[width=1in,height=1.25in,clip,keepaspectratio]{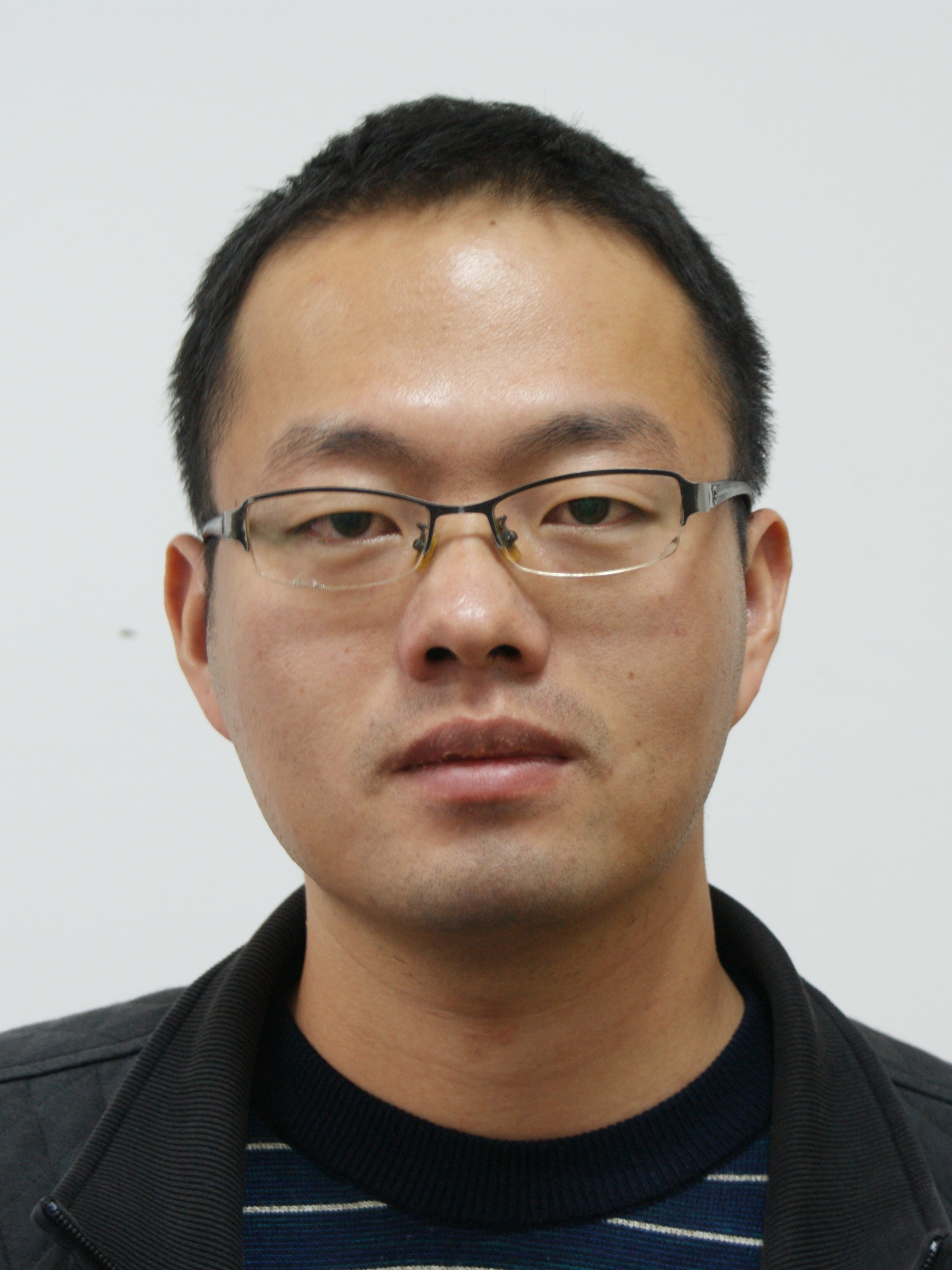}}]{Liguo Zhou}
received a bachelor's degree in software engineering from Suzhou University, a master's degree in pattern recognition and intelligent systems from Wuhan University, and a doctoral degree in computer science from the Technical University of Munich. He conducts research in artificial intelligence at Huaibei Normal University, with interests in computer vision, autonomous driving, and embodied intelligence.
\end{IEEEbiography}

\vspace{11pt}

\begin{IEEEbiography}[{\includegraphics[width=1in,height=1.25in,clip,keepaspectratio]{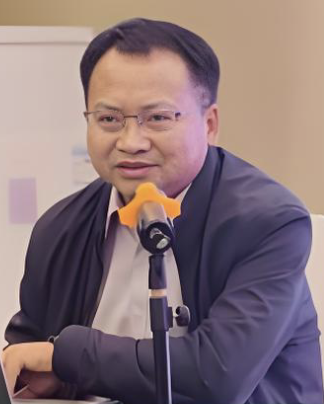}}]{Lei Yu}
is a professor in the School of Computer Science and Technology at Huaibei Normal University. His research interests include artificial intelligence, information security, and security protocols.
\end{IEEEbiography}

\vspace{11pt}

\begin{IEEEbiography}[{\includegraphics[width=1in,height=1.25in,clip,keepaspectratio]{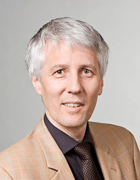}}]{Alois Knoll}
(Fellow, IEEE) received a diploma in electrical engineering from the University of Stuttgart in 1985 and a doctorate in computer science from the Technical University of Berlin in 1988. He was a professor at Bielefeld University from 1993 to 2001 and has been a professor at the Technical University of Munich since 2001. His research focuses on autonomous systems, robotics, and artificial intelligence.
\end{IEEEbiography}